\documentclass{article}

\PassOptionsToPackage{numbers, sort&compress}{natbib}

\usepackage[final]{neurips_2025}

\usepackage{pifont}
\usepackage{subfigure}

\usepackage[utf8]{inputenc} % allow utf-8 input
\usepackage[T1]{fontenc}    % use 8-bit T1 fonts
\usepackage[dvipsnames]{xcolor}
\usepackage[colorlinks=true,
            linkcolor=BrickRed,     % for cross-references like \ref, \autoref
            citecolor=teal,     % for citations like \cite
            urlcolor=blue    % for \url and \href
]{hyperref}
\usepackage{url}            % simple URL typesetting
\usepackage{booktabs}       % professional-quality tables
\usepackage{amsfonts}       % blackboard math symbols
\usepackage{nicefrac}       % compact symbols for 1/2, etc.
\usepackage{microtype}      % microtypography
\usepackage{xcolor}         % colors
\usepackage{graphicx} 
\usepackage{amsmath}
\usepackage{amssymb}
\usepackage{wrapfig}
\usepackage{multirow} 
\usepackage{dsfont}
\usepackage{enumitem} % add this in your preamble
\usepackage{bm}
\usepackage{authblk} % For handling affiliations

\newcommand{\revisions}[1]{#1}

\usepackage{xspace}
\newcommand{\spikachump}{\texttt{Spikachu-mp}\xspace}

\makeatletter
\renewcommand\AB@affilsepx{, \protect\Affilfont}
\makeatother

\title{A Scalable, Causal, and Energy Efficient Framework for Neural Decoding with Spiking Neural Networks}

\author[1, *]{Georgios Mentzelopoulos}
\author[1, *]{Ioannis Asmanis}
\author[1]{Konrad P. Kording}
\author[1]{Eva L. Dyer}
\author[1, 2, \textdagger]{Kostas Daniilidis}
\author[1, \textdagger]{Flavia Vitale}
\affil[1]{University of Pennsylvania}
\affil[2]{Archimedes, Athena RC}

\begin{document}

\renewcommand{\thefootnote}{\fnsymbol{footnote}}

\footnotetext[1]{Equal contribution. The order of the co-first authors was determined by a coin flip and is fully interchangeable. Contact: \{gment, asmanis\}@seas.upenn.edu}

\footnotetext[2]{Equal contribution. Contact: \{kostas@cis, vitalef@pennmedicine\}.upenn.edu, \\ Project page and code: \href{https://spikachu-bci.github.io}{\texttt{https://spikachu-bci.github.io}}}

\maketitle

\begin{abstract}
Brain-computer interfaces (BCIs) promise to enable vital functions, such as speech and prosthetic control, for individuals with neuromotor impairments.
Central to their success are neural decoders, models that map neural activity to intended behavior.  
Current learning-based decoding approaches fall into two classes: simple, causal models that lack generalization, or complex, non-causal models that generalize and scale offline but struggle in real-time settings. 
Both face a common challenge, their reliance on power-hungry artificial neural network backbones, which makes integration into real-world, resource-limited systems difficult.
Spiking neural networks (SNNs) offer a promising alternative. 
Because they operate causally (i.e. only on present and past inputs) these models are suitable for real-time use, and their low energy demands make them ideal for battery-constrained environments.
To this end, we introduce \emph{Spikachu: a scalable, causal, and energy-efficient neural decoding framework based on SNNs}. 
Our approach processes binned spikes directly by projecting them into a shared latent space, where spiking modules, adapted to the timing of the input, extract relevant features; these latent representations are then integrated and decoded to generate behavioral predictions.
We evaluate our approach on 113 recording sessions from 6 non-human primates, totaling 43 hours of recordings. 
Our method outperforms causal baselines when trained on single sessions using between 2.26× and 418.81× less energy.
Furthermore, we demonstrate that scaling up training to multiple sessions and subjects improves performance and enables few-shot transfer to unseen sessions, subjects, and tasks. 
Overall, Spikachu introduces a scalable, online-compatible neural decoding framework based on SNNs, whose performance is competitive relative to state-of-the-art models while consuming orders of magnitude less energy.

\end{abstract}

\section{Introduction}
\label{sec:intro}
Brain-computer interfaces (BCIs) are opening new frontiers in assistive technology, particularly for those affected by severe neuromotor disorders \cite{Lorach2023, Cole2020, SingerClark2025, Xu2022, Zimmerman2024, Kim2023, Kim2023A175, Mentzelopoulos2023, Erickson2024tms, Willett2021text2speech}.
The implantation of miniaturized BCI devices in the brains of patients suffering from debilitating disorders, such as limb loss or ALS, enabled them to control 
computers and smartphones using only their thoughts, vastly improving their quality of life \cite{nuyujukian2018cortical, Hettick2022, Oxley2022}.
\revisions{Latest advances have even restored speech-based communication in individuals who had lost the ability to speak by reconstructing their intended words directly from neural activity \citep{Willett2023, Metzger2023}}.
At the core of every BCI system lies a neural decoder, the software that maps neural signals
to the user's intended actions, such as moving a cursor on a screen or controlling a prosthetic limb \cite{Sajda2008, Collinger2013, Flesher2021, Willsey2025}. 

Deep learning has significantly advanced our ability to build powerful neural decoders \cite{Metzger2023, azaboumulti, Willett2023, Schneider2023}. 
Unlike traditional approaches that rely on hand-crafted features, artificial neural networks (ANNs) can learn the mapping from neural activity to intended actions directly from data. 
Despite this progress, designing scalable, high-performance neural decoders suitable for integration into real-world BCI systems remains an open challenge with constraints along several axes. 
First, neural decoders need to be energy-efficient to operate within the tight power budgets of battery-constrained implantable devices \citep{Sahasrabudhe2023, Driscoll2024, Erickson2024Evaluating, Mitchell2023}. Second, to enable online operation, models must be causal (i.e. rely only on present and past inputs) \citep{Willett2021text2speech}.
Third, models need to scale, since scaling up model complexity and training dataset size has been shown to boost performance in the neural decoding domain \citep{azaboumulti, Schneider2023, Azabou2023, mentzelopoulos2024neural, Wang2025foundation}. 
Finally, models should generalize to new subjects and tasks with minimal training examples to reduce the need for lengthy calibration sessions that hinder the practical deployment of BCIs \cite{Kording2024neural}.

While significant progress has been made along each of these axes, to our knowledge, no single framework excels across all of them simultaneously. 
Existing approaches tend to fall into two main categories, each with its own shortcomings. On one hand, simple models, often based on traditional architectures like multi-layer perceptrons (MLPs) or Gated Recurrent Units (GRUs), tend to perform well within individual experiments. These methods are typically causal but require 
homogeneous input structures, making them difficult to scale or generalize across subjects \citep{Serruya2002, Ethier2012, ODoherty2011, Jarosiewicz2015, Willett2019}. On the other hand, more sophisticated frameworks that can be trained across datasets have demonstrated strong performance and generalization, particularly at scale \citep{azaboumulti, Schneider2023, Azabou2023, mentzelopoulos2024neural, Wang2025foundation, Sussillo2016}. However, their lack of causal processing and heavy computational demands challenges their applicability outside the research lab. %Models from both categories also face a common challenge, they rely on power-hungry artificial neural network (ANN) backbones, which makes integration into real-world, resource-limited systems difficult. 

Spiking neural networks (SNNs) offer a promising alternative. Their inherent causality supports integration into online systems, and their low computational footprint makes them well-suited for battery-constrained environments.
To this end, we introduce \emph{Spikachu: a scalable, causal, and energy-efficient framework for multi-session, multi-subject neural decoding based on SNNs}. %Our method combines a causal ``harmonization'' module, inspired by the POYO encoder \citep{Azabou2023}, followed by a processing module, based on SNNs%, which performs the bulk of computation to get behavioral predictions very efficiently. 
Our approach operates on binned spike trains, which are first projected into a latent space shared across sessions and subjects. Temporal features are then extracted from the latents using parallel spiking networks. To capture long-range dependencies, the extracted features are processed by spiking self-attention blocks. Finally, the enriched latents are integrated and mapped to behavioral predictions through another set of parallel spiking modules.

We evaluate our approach on 113 neural recording sessions from 6 non-human primates (NHPs) totaling more than 111M spikes and 43 hours of recordings \citep{Perich2018, Pei2021nlb}.
Our approach outperforms causal baselines while consuming between 2.26× and 418.81× less energy when trained on single sessions. 
We then build unified models trained on multi-subject data that outperform single-session models, and show they can be transferred to new sessions, subjects, and tasks very efficiently.
Overall, this work introduces a scalable, online-compatible neural decoding framework based on SNNs, whose performance is competitive relative to state-of-the-art models while consuming orders of magnitude less energy. This combination of performance and efficiency makes it a promising foundation for BCIs designed for edge computing environments.

Our contributions can be summarized as:
\begin{itemize}[itemsep=0.4pt, topsep=1pt, parsep=0.4pt, partopsep=.40pt]
\item \emph{A framework for multi-session, multi-subject neural decoding that is scalable, causal, and energy efficient.} By combining SNNs with transformers, our approach offers strong decoding performance while using minimal energy, positioning it as a compelling solution for real-time BCI applications.
\item \emph{A causal, architecture agnostic building block for multi-session, multi-subject model training.} We propose a novel formulation for mapping neural recordings from heterogeneous datasets into a shared latent space, enabling scalable training across sessions and subjects. % despite the lack of correspondence of the neural activity across datasets.
\item \revisions{
    \emph{A fast, architecture agnostic building block for efficient processing across temporal resolutions}. We propose a novel module capable of extracting features at multiple, distinct temporal scales based on SNNs.
} 
\item \emph{Pretrained models for neural decoding.} We trained a unified model 
on the combined data from 3 NHPs spanning 99 neural recording sessions, 
which is transferable to new sessions, subjects, and tasks. We will make the model and code publicly available. %as a resource for the community.
\end{itemize}

\section{Related Work}

\paragraph{Neural decoding.}

Recent advances in deep learning have significantly improved neural decoding capabilities, with ANNs enabling impressive results across a range of applications, including cursor control, prosthetic control, typing, and speech decoding \citep{Metzger2023, Willett2021text2speech, Collinger2013, Flesher2021, Willsey2025, Willett2023, Willett2019}. 
These works primarily relied on lightweight, causal architectures such as MLPs and RNNs, which performed well on single-session datasets \cite{Glaser2020}. \revisions{Relatively more sophisticated approaches based on VAEs and self-supervision have also been used with success by \citet{liu2021drop} and \citet{peterson2022learning}, respectively.} 
However, their performance degrades on new sessions or subjects due to their reliance on homogeneous input structures and known electrode correspondences across sessions and subjects.

Other studies have focused on improving cross-session generalization. 
LFADS was among the first to model latent dynamics across sessions and subjects \citep{Sussillo2016}, with extensions for large-scale training \citep{keshtkaran2022large}. 
CEBRA introduced a contrastive learning approach for learning neural representations that are shared across subjects \citep{Schneider2023}, while POYO used transformers to achieve multi-session, multi-subject, and multi-task generalization \citep{Azabou2023}. 
POYO+ extended those results across distinct cell types and brain regions \citep{azaboumulti}. 
\revisions{The MICrONS foundation model demonstrated that large-scale pretraining on the mouse visual cortex can yield neural representations that generalize across visual stimuli \citep{Wang2025foundation}}.
While these approaches perform impressively and generalize well, they often sacrifice causality and introduce significant computational overhead, which is unsuitable for online applications such as battery-constrained BCIs.

\paragraph{Spiking neural networks.}

SNNs are a class of neural networks that excel in processing event-driven data sequences with spikes \citep{Malcolm2023snn}.
Unlike conventional ANNs, which rely on static activations like ReLU, SNNs employ bio-inspired spiking mechanisms, with the Leaky Integrate-and-Fire (LIF) model being the most widely used (see App.~\ref{app:snns}). Key to their success is their remarkably energy-efficient inference compared to ANNs, particularly when deployed on neuromorphic hardware \citep{orchard2021, true-north, Davies2018Loihi}. Furthermore, their event-driven, online nature makes them well-suited for processing real-time or asynchronous data streams \citep{spikeflownet, decroon-multiscale, xu-evb-flow-sparsification, zhu2024autonomous}.

SNNs have been investigated for their potential in energy-efficient neural decoding \citep{Dethier2011, Dethier2013, Liao2022, Leone2023, Taeckens2023}. However, prior efforts have primarily focused on shallow spiking MLPs, typically with four or fewer layers, trained and evaluated within the confines of single recording sessions. The main emphasis of these works has been on deployment feasibility on neuromorphic hardware.
In contrast, our work introduces the first SNN-based neural decoding framework that not only leverages the inherent energy efficiency of spiking computation but also scales to large, multi-session, and multi-subject datasets, marking a significant step forward in the applicability of SNNs to real-world neural decoding challenges.

\section{Methodology}

SNNs offer remarkable energy efficiency across a variety of tasks in computer vision and natural language processing \citep{Malcolm2023snn, zhou2023spikformer, spikingbert}, particularly with asynchronous, event-based data \citep{decroon-multiscale}. Our work aims to leverage their advantages in the context of neural decoding.

\begin{figure}[h]
    \centering
    \includegraphics[scale=0.55]{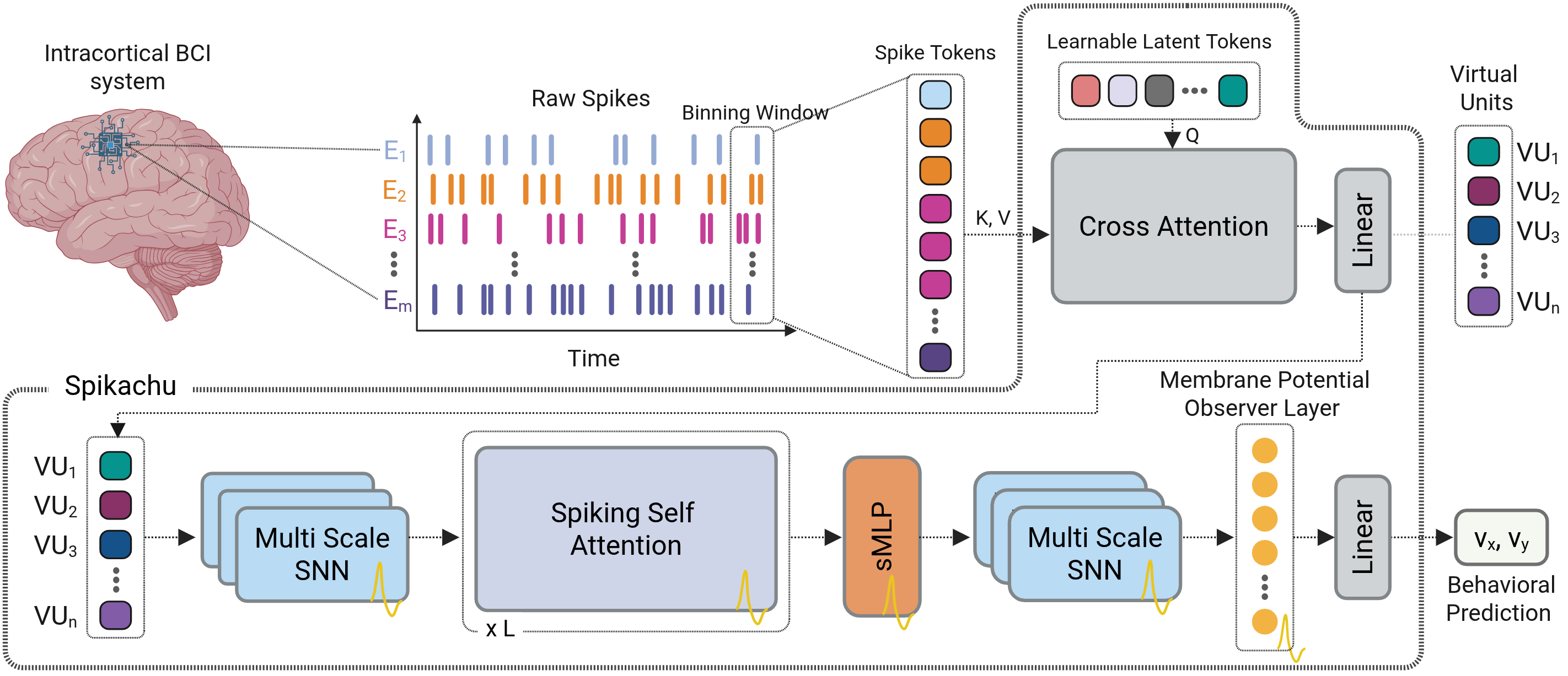} 
    \caption{\textit{Overview of the Spikachu framework}.} %Spikes are binned using a rolling window and project to a latent space using the \incomplete{POYO/PerceiverIO} encoder. Feature patches are extracted via a set of parallel spiking MLPs, which are processed with sequential spiking self-attention blocks. The latents are mixed and projected though another set of spiking MLPs. The latents are projected through a final membrane potential readout layer and a linear layer to return behavioral predictions.}
    \label{networkArchitectrure}
\end{figure}

\subsection{Harmonizing the neural activity across sessions and subjects} \label{sec:harmonizer}

BCI systems rely on microelectrode arrays (MEAs) to record neural activity at a high spatial and temporal resolution. Thanks to their small size and low impedance, individual MEA electrodes can detect extracellular action potentials from single neurons or small groups of neurons (referred to as ``units'') in their immediate vicinity \citep{Obien2015, MXtrodepaper}. These recordings are typically represented as spike trains, which abstract away the waveform of each action potential and encode only the precise times at which spikes occur.

The spikes recorded across the electrodes of a subject are, of course, not  independent. Rather, they reflect the coordinated activity of neurons that are part of distributed brain networks \citep{Azabou2023}.
A key challenge is interpreting these signals not as isolated events, but as components of a broader neural context \citep{mentzelopoulos2024neural}. 
This challenge becomes increasingly complex when integrating data across multiple sessions and subjects. Within a single subject, MEA recordings can drift over time  
\citep{Hofmann2006, Obien2015, Azabou2023, ODoherty2011}. 
Across subjects, MEAs inevitably sample from distinct neurons, with no known correspondence between electrodes \citep{dabagia2023aligning}. Consequently, our decoding approach must be %robust to this variability 
capable of extracting meaningful, generalizable representations despite differences in electrode placement, neural dynamics, and recorded populations across sessions and individuals.

\paragraph{Tokenization.} Motivated by the heterogeneity of the neural recordings across sessions and subjects, we designed a tokenization scheme that enables multi-session training.
At the same time, our approach operates within timepoints to enable online usability.
Inspired by \citet{Azabou2023}, we represent each unit as a token via a learnable embedding in $\mathbb{R}^{d}$.
Specifically, for a binned spike window,
let
$\mathcal{U} = \{u_{1}, u_{2}, u_{3}, \ldots, u_{n_{u}}\}$
be the multiset of all units recorded during that window.
Let UnitEmb($\cdot$) denote a lookup table that maps each unit to its embedding.
We summarize the neural activity of a given time window as the sequence
$\mathbf{X} = [ \mathbf{x}_{1}, \mathbf{x}_{2}, \mathbf{x}_{3}, \dots, \mathbf{x}_{n_{u}} ] \in \mathbb{R}^{n_{u} \times d}$,
where $\mathbf{x}_{i} = \text{UnitEmb(}u_{i}{)} ~ \forall ~ i \in \{1,2, \cdots, n_{u}\}$.
\revisions{We note that if the $i^{th}$ electrode recorded $n_{i}$ units, $n_{i}$ repeats of $ \mathbf{x_{i}} = \text{UnitEmb(}u_{i}{)}$ would be present in the sequence $\mathbf{X}$ for that window.}
This sequence summarizes the cumulative neural activity across all electrodes of a subject for the given window.

\paragraph{Projecting units to a shared latent space.} We then project the sequence $\mathbf{X}$ into a latent space that is shared across sessions and subjects using the Perceiver encoder \citep{Jiagle2021Perceiver, Jeagle2021}. Specifically, let 
$\mathbf{Z}_{0} = [ \mathbf{z}_{0, 1}, \mathbf{z}_{0, 2}, \cdots, \mathbf{z}_{0, n_{0}}] \in \mathbb{R}^{n_{0} \times d_{z_{0}}}$,
be a sequence of $n_{0}$ learnable latent tokens, where $\mathbf{z}_{0, i} \in \mathbb{R}^{d_{z_{0}}} ~ \forall ~ i \in \{1, 2, \dots, n_{0}\}$. 
We use cross-attention to project the input sequence $\textbf{X}$ into the latent $\mathbf{Z}_{1} \in \mathbb{R}^{n_{0} \times d}$ whose length $n_{o}$ is independent of the length of the input sequence $\mathbf{X}$. 
To do so, we linearly project the latent and input sequences into 
$\text{Queries: }\mathbf{Q} = \mathbf{W}_{Q} \mathbf{Z}_{0}, ~\text{Keys: }\mathbf{K} = \mathbf{W}_{K} \mathbf{X}, ~ \text{and Values: } \mathbf{V} = \mathbf{W}_{V} \mathbf{X}$,
and compute,
\begin{equation}
    \mathbf{Z}_{1} \leftarrow \text{Cross-Attn}(\mathbf{Q}, \mathbf{K}, \mathbf{V}) = \text{Softmax}\left(
        \frac{\mathbf{QK}^{\top}}{\sqrt{d_{k}}}
    \right) \mathbf{V},
    \label{eq:attn}
\end{equation}
where $d_{k}$ refers to the feature dimensionality of $\mathbf{K}$. For this operation, we use the standard transformer block preceded by layer-normalization and followed by a feed-forward network \citep{Vaswani2017}. 

\paragraph{Obtaining virtual units.} We then unroll the latent $\mathbf{Z}_{1} \in \mathbb{R}^{n_{0} \times d} \rightarrow \mathbb{R}^{n_{0} \cdot d}$ and we project it to the low-dimensional latent $\mathbf{Z}_{2} \in \mathbb{R}^{n_{v}}$ using a linear layer. We interpret $\mathbf{Z}_{2}$ as a new set of $n_{v}$``virtual'' units that are shared across sessions and subjects. We note that the weights of this layer are shared across sessions and subjects---no session or subject specific adaptation is performed.

\subsection{Efficient and stateful processing of the latents using spiking neural networks}
\label{sec:snn-processing}

Having projected the neural activity for each time bin into a new set of virtual units that is shared across sessions and subjects, we could in principle process the latent $\mathbf{Z}_{2}$ with any causal architecture commonly used for neural decoding such as MLP, GRU, or any model surveyed by \citet{Glaser2020}. 

Rather than relying on such conventional power-hungry ANNs, we instead leverage SNNs which are considered the third generation of neural network models \citep{Maass1997spiking}. SNNs are theoretically as expressive as ANNs and, due to their recurrent nature, are particularly well-suited for time-dependent tasks \citep{Boahen2017neuromorphic}. While their performance advantages remain an open area of research, their energy efficiency when deployed on neuromorphic hardware has been well demonstrated \citep{true-north, Davies2018Loihi}. This makes SNNs particularly attractive for BCI applications, where power constraints are critical due to the limited battery life of the miniaturized BCI implants.

\paragraph{Decomposing the neural activity across temporal streams.}
Given that the brain operates across multiple intrinsic timescales \citep{Murray2014timescales}, we sought to process the data in a similarly multi-timescale fashion. 
To do this, we pass the latent $\mathbf{Z}_{2}$ through $p$ parallel spiking feed forward networks (FFNs), each designed to operate at a distinct temporal resolution. 
Each network returns a latent $\mathbf{Z}_{s,i} \in \mathbb{R}^{d_{s}}$ by projecting $\mathbf{Z}_{2}$ though sequences of the following layers,
\begin{equation}
    \mathbf{Z}_{l+1} = \textrm{BN}(\mathbf{W}_{l} ~\mathcal{SN}_{l}(\mathbf{Z}_{l})),
    \label{spikingffn}
\end{equation}
where $\mathcal{SN}_{l}$ represents a spiking activation, $\mathbf{W}_{l}$ a linear projection, and BN the batch normalization operation \citep{Ioffe2015BatchNorm}.
Each network's spiking activation layers $\mathcal{SN}$ are initialized with independent and learnable decay constants (see App. \ref{app:snns}), effectively allowing each network to extract features that evolve as separate streams of information. We concatenate the latents from each stream into the sequence of latent tokens $\mathbf{Z}_\textrm{ms} = \text{[}\mathbf{Z}_{s,1}, \mathbf{Z}_{s, 2}, \cdots, \mathbf{Z}_{s,p}\text{]} \in \mathbb{R}^{p \times d_{s}}$.

\paragraph{Capturing long-range temporal dependencies.} We then proccess the sequence $\mathbf{Z}_\textrm{ms}$ for informational dependencies across timescales using spiking-self attention \citep{zhou2023spikformer}. 
Specifically, we project the sequence $\mathbf{Z}_\textrm{ms}$ into equally shaped Queries ($\mathbf{Q}$), Keys ($\mathbf{K}$), and Values ($\mathbf{V}$),
\begin{equation}
    \mathbf{Q} = \mathcal{SN}_{Q} (\textrm{BN} (\mathbf{W}_{Q}\mathbf{Z}_{\textrm{ms}})), ~~~~ \mathbf{K} = \mathcal{SN}_{K} (\textrm{BN} (\mathbf{W}_{K} \mathbf{Z}_{\textrm{ms}})), ~~~~ \mathbf{V} = \mathcal{SN}_{V} (\textrm{BN} (\mathbf{W}_{V}\mathbf{Z}_{\textrm{ms}})),
\end{equation}
where $\mathbf{Q, ~K, ~V} \in \mathbb{R}^{d_\textrm{ssa}}$ and compute the spiking self attention matrix as,
\begin{equation}
    \text{SSA}^\prime = \mathcal{SN}(\mathbf{QK}^\top \mathbf{V} \cdot s),
\end{equation}
where $s$ is a scaling factor. %(note that no softmax activation is used).
We then compute $\mathbf{Z_\textrm{ssa}} \in \mathbb{R}^{p \times d_\textrm{ssa}}$ by projecting SSA$^\prime$ through the following spiking layers,
\begin{equation}
    \mathbf{Z}_{\textrm{ssa}} \leftarrow \textrm{SSA}(\mathbf{Q}, \mathbf{K}, \mathbf{V}) = \mathcal{SN}(\textrm{BN}(\mathbf{W}_{\textrm{ssa}}~ \textrm{SSA}^{\prime})).
\end{equation}
Following \citet{zhou2023spikformer}, we precede this operation using layer normalization and follow it with a spiking feed-forward network. For a primer on the main differences between spiking \citep{zhou2023spikformer} and vanilla \citep{Vaswani2017} self-attention, please see App. \ref{sec:vsa_vs_ssa}.

\paragraph{Compressing the latents to a compact spatiotemporal representation.}
Having identified long-term dependencies across the $p$ temporal streams, we unroll the latents $\mathbf{\mathbf{Z}}_\textrm{ssa} \in \mathbb{R}^{p \times d_\textrm{ssa}} \rightarrow \mathbb{R}^{p \cdot d_\textrm{ssa}}$ and using a spiking MLP (whose layers follow Eq. \ref{spikingffn}), we project them to a low dimensional representation $\mathbf{Z}_\textrm{mlp} \in \mathbb{R}^{d_\textrm{mlp}}$ where $ d_\textrm{mlp} \ll p \cdot d_\textrm{ssa}$ .
The latent $\mathbf{Z}_\textrm{mlp}$ is a compact representation of the neural activity that is informed by the evolution of the neural code along the spatial (electrode) and temporal axes. 

\paragraph{Smoothing predictions using multiple timescales.}
Having extracted neural representations that capture long-range spatiotemporal dependencies, we process the latent $\mathbf{Z}_\textrm{mlp}$ with another set of $p'$ parallel spiking MLPs that process the latents at different temporal scales (composed of layers described in Eq. \ref{spikingffn}). This time, we use the parallel MLPs to extract latents $\mathbf{Z}_{s', i} \in \mathbb{R}^{d_{s'}}$, each evolving at unique temporal scales based on the decay constant of the neurons in that network. We then concatenate the latents to obtain the sequence $\mathbf{Z}_{\textrm{sm}'} \in \mathbb{R}^{p' \times d_{s'}}$. %\incomplete{Those latents, when combined linearly can ``smoothly'' change the membrane potential of the subsequent layer, which we use to track our behavioral variables of interest. }

\paragraph{Tracking continuous variables through the membrane potential of spiking neurons.}

In many BCI applications, the behavioral variables of interest, such as the velocity of a computer cursor or the movement of a prosthetic limb, are continuous in nature. To accommodate this, our framework needs to map the spiking activity generated by the SNN layers to continuous-valued outputs. We achieve this using a
membrane potential observer layer, denoted $\overline{\mathcal{SN}}_{\textrm{obs}}$, that never spikes and whose membrane potential fluctuates over time without resetting \citep{Liao2022, Shaikh2019, spiking-decoder}. We first unroll the latents $\mathbf{Z}_{\textrm{sm}^\prime} \in \mathbb{R}^{p' \times d_{s'}} \rightarrow \mathbb{R}^{p' \cdot d_{s'}}$ and then pass them through the observer layer,
\begin{equation} \mathbf{M}_{\textrm{obs}} \leftarrow \overline{\mathcal{SN}}_{\textrm{obs}}(\mathbf{Z}_{\textrm{sm}^{\prime}}), \end{equation}
where $\mathbf{M}_{\textrm{obs}} \in \mathbb{R}^{p' \cdot d{s'}}$ denotes the membrane potential of the observer neurons. These accumulated latents are then linearly projected to the target variable's output space %via a linear projection
producing $\mathbf{Z}_{\textrm{out}} \in \mathbb{R}^{d_{\textrm{out}}}$, the final continuous behavioral prediction of the network.

\section{Experiments} \label{sec:experiments}

In this section, we validate the promise of our approach (see Fig. \ref{networkArchitectrure}) for causal, energy-efficient and scalable neural decoding based on spiking neural networks.

\subsection{Experimental setup}

\textbf{Dataset.} We utilized two publicly available electrophysiology datasets summarized in Tab. \ref{tab:datasets}. 
Altogether, the data span 6 NHPs engaged in 4 distinct behavioral tasks, encompassing a total of 113 recording sessions \citep{Perich2018,perich2017altered,glaser2018population,gallego2020long, Pei2021nlb}. 
In terms of scale, \emph{the combined dataset contains over 43 hours of recordings, 10,000 units, 110 million spikes, and 20 million behavioral timepoints}, providing a rich and diverse foundation for evaluating the scalability of our approach.
\begin{table} [h]
\centering
\begin{caption}{ \textit{Datasets used in this work.  CO: Center-Out, RT: Random Target.}} \label{tab:datasets}
\end{caption}
\begin{tabular}{l|p{1.2cm}rrrrrp{1.2cm}r}
\hline
\footnotesize{{\bf Study}} & \footnotesize{{\bf Regions}} & \footnotesize{{\bf \# Indiv}} & \footnotesize{{\bf \# Sess}} & \footnotesize{{\bf \# Units}} & \footnotesize{{\bf \# In }} & \footnotesize{{\bf \# Out}} & \footnotesize{{\bf Tasks}} \\ \hline
\footnotesize{\citet{Perich2018}} & \footnotesize{M1, PMd} & 4  & 111         & 10,410    & 111.39M            & 20M              &  \footnotesize{CO, RT}     \\  % 

 \footnotesize{\citet{Pei2021nlb}}     & \footnotesize{M1}     & 2           & 2          & 312    & 5M            & 9M              & \footnotesize{RT, Maze} \\  %  \cite{pei2021nlb}

 \hline
\end{tabular}

\vspace{-5mm}
\end{table}

\paragraph{Behavioral tasks.} 
Neural recordings were collected from NHPs that performed motor tasks of various complexities (see Fig. \ref{fig:single_session_training}A and App. \ref{appendix:sec:data}). In the CO task, the animal executes a relatively structured sequence: after receiving a go cue, it reaches toward one of eight predefined targets before returning to the center.  The RT task presents additional complexity. The animal engages in continuous, self-paced movements, with new targets appearing unpredictably across the workspace. 
\emph{Our goal was to decode the velocity of the cursor controlled by the animal from their neural activity}. 

\paragraph{Design choices.} 
Throughout all experiments, we bin spikes using a 0.01 sec sliding window on 1 sec segments of data. 
\revisions{We do not use the trial structure during model training.} 
We report the decoding performance for CO tasks during reaching movements only, as established in \citet{Azabou2023}. Spikachu's implementation and training details are described in App. \ref{app:model_implementation_details} and \ref{app:spikachu-training}.

\paragraph{Energy estimation.}

\revisions{
All energy-related results in this work are derived from the number of floating-point operations (FLOPs) required for model inference, which we convert to energy consumption estimates following well-established procedures described in  \citet{spikingbert} and \citet{zhu2024autonomous}. \emph{Because FLOPs are hardware-agnostic, our analyses do not depend on any specific hardware implementation}. All details of the energy calculations are provided in App. \ref{app:energy-snn}.
}

\subsection{Performance on single sessions}
\label{sec:ss_training}

\begin{figure}[h]
    \centering
    \includegraphics[scale=0.42]{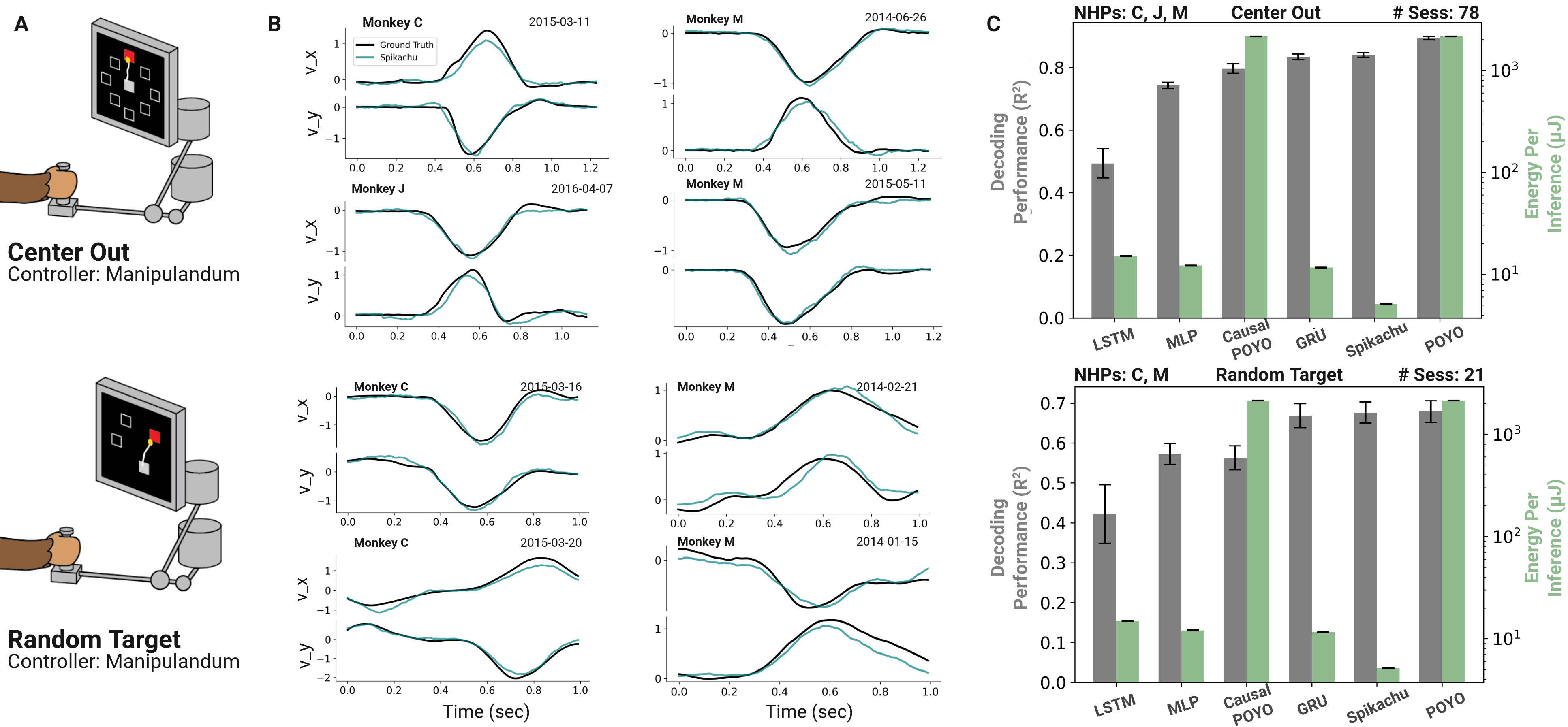} 
    \caption{\textit{Model performance on single sessions}. \textbf{(A)} Schematic of CO and RT tasks (reproduced from \citet{Azabou2023}, with permission). \textbf{(B)} Examples of true and predicted behavior (x, y velocities) for the CO (top 4 subplots) and RT (bottom 4 subplots) tasks. \textbf{(C)} Mean ($\pm$ SEM) decoding performance and energy consumption for Spikachu and baselines on sessions from monkeys C, J, and M from the \citet{Perich2018} dataset.}
    \label{fig:single_session_training}
\end{figure}

We began by evaluating the performance of our modeling approach in a single-session setting. Specifically, we trained individual models on 99 recording sessions from three animals (NHPs C, J, and M) from the \citet{Perich2018} dataset. Each session ranged from 10 to 106 minutes in duration, comprising 78 sessions of the CO task and 21 sessions of the RT task. 
These models achieved an \textit{average per-session $R^{2}$ of 0.84 and  0.68 
for the CO and RT tasks}, respectively (see Fig. \ref{fig:single_session_training}B). We also estimated the mean energy required per inference across the single-session models (see App. \ref{app:energy-snn} for details) and found it to be $5.14 \mu J$ and $5.13\mu J$ for the CO task and 
RT tasks, respectively.

\begin{wraptable}{r}{0.6\textwidth}
\vspace{-4mm}
\centering
\caption{\footnotesize{\textit{Model performance for Spikachu and baselines}. Best performing model is in bold and second best model is underlined.}}
\footnotesize
\begin{tabular}{l|rr|rr}
\hline
\multirow{2}{*}{\textbf{Model}} 
& \multicolumn{2}{c|}{\textbf{Decod. Perf. (R\textsuperscript{2}) $\uparrow$}} 
& \multicolumn{2}{c}{\textbf{Energy (\bm{$\mu$}J) $\downarrow$}} \\
& \multicolumn{1}{c}{\textbf{CO}} & \multicolumn{1}{c|}{\textbf{RT}} 
& \multicolumn{1}{c}{\textbf{CO}} & \multicolumn{1}{c}{\textbf{RT}} \\
\hline
LSTM & 0.4935 & 0.4214 & 15.08 & 14.94 \\
MLP & 0.7424 & 0.5724 & 12.18 & 12.06 \\
POYO-causal & 0.7961 & 0.5629 & 2151.65 & 2136.82 \\
GRU & 0.8336 & 0.6681 & \underline{11.65} & \underline{11.54} \\
POYO & \textbf{0.8937} & \textbf{0.6785} & 2151.65 & 2136.82 \\

Spikachu  & \underline{0.8398} & \underline{0.6761} & \textbf{5.14} & \textbf{5.13} \\
\hline
\end{tabular}

\label{tab:baseline_comprisons}
\vspace{-3mm}
\end{wraptable}

\paragraph{Baseline comparisons.} We then benchmarked the single session models against other architectures commonly used for neural decoding, such as MLP, GRU, and POYO \citep{Azabou2023, Glaser2020}. Our approach outperformed all causal baselines 
(see Fig. \ref{fig:single_session_training}C and Tab. \ref{tab:baseline_comprisons}),
while narrowing the performance gap with non-causal models. \textit{Notably, Spikachu was the most energy-efficient model, requiring 2.26× less energy per inference when compared to GRU (the second most efficient model) and 418.81× less when compared to POYO, which was the best performing model}. 
\revisions{Importantly, the same trend was observed when comparing the number of FLOPs required for inference for each model (see Tab. \ref{tab:flops}, App. \ref{sec:memory_ops})}. 
These findings highlight the dual benefits of our approach: high-performance neural decoding and improved energy efficiency, underscoring its suitability for power-constrained applications such as implantable BCIs. 

\revisions{We also benchmarked our model against baselines in terms of memory access costs (where Spikachu outperformed all baseline models). We refer the reader to App. \ref{sec:memory_ops} for this analysis. Detailed implementation descriptions of the baseline models and the corresponding energy calculations are provided in App. \ref{app:baselines} and App. \ref{app:energy-baselines}, respectively.}

\subsection{\spikachump: Pretraining on large amounts of data}
\label{sec:ms_training}
After demonstrating Spikachu's strong performance when trained on individual sessions, we were interested in investigating whether training on more data, despite the heterogeneity, could further enhance Spikachu's performance, a strategy that has shown benefits in prior works \cite{Azabou2023, mentzelopoulos2024neural, Schneider2023, Wang2025foundation}. 
\begin{wrapfigure}{r}{0.55\textwidth} 
    \centering
    \includegraphics[scale=0.45]{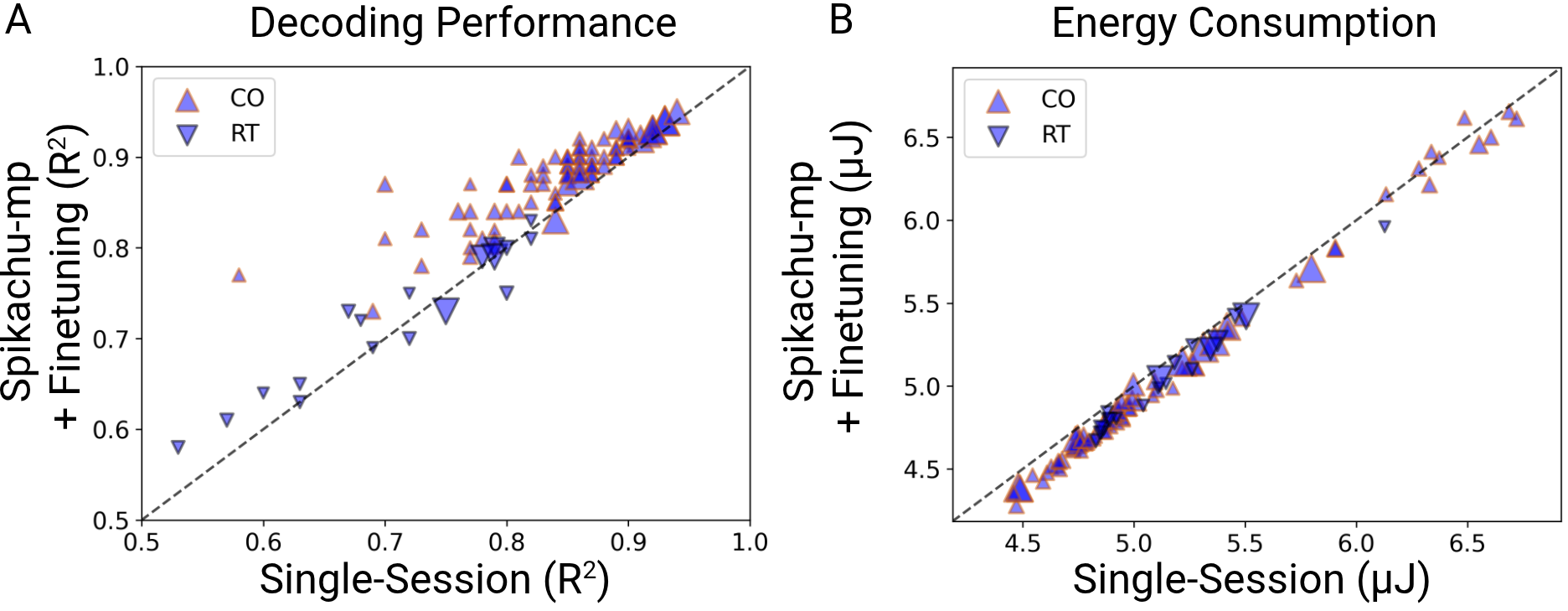}
    \caption{\textit{Head-to-head comparison between \spikachump + finetuning vs Spikachu trained on single-sessions.} \textbf{(A)} Decoding Performance, \textbf{(B)} Energy consumption per inference. Marker size encodes the number of trials available  for each session.}
    \label{fig:ss_vs_msf}
\end{wrapfigure}
To this end, we developed \spikachump: a multi-session, multi-subject model trained on the complete set of 99 recording sessions from monkeys C, J, and M from the \citet{Perich2018} dataset. \emph{Notably, this model was trained on over 40 hours of neural recordings and more than 110 million spikes.} 
\revisions{In contrast to previous works (see \citet{Azabou2023}), we did not increase our model's size during multi-subject training. This design choice ensured that our model would remain energy-efficient even when scaling to larger datasets, a critical consideration for resource-constrained implantable BCIs.}
\spikachump achieved a per-session test set $R^{2}$ of 0.80  for CO and 0.57 for the RT task,
indicating Spikachu's ability to learn generalized neural representations shared across sessions and subjects.

To evaluate the utility of the learned representations, we then finetuned \spikachump on individual sessions. The finetuned models achieved an average test set $R^{2}$ of 0.88 and 0.69
for the CO and RT tasks, respectively, outperforming models trained from scratch on single sessions ($\Delta R^{2}_{\textrm{CO}} = 3.75$\% and $\Delta R^{2}_{\textrm{RT}} = 1.71$\%; Fig.~\ref{fig:ss_vs_msf}A). 
The finetuned models also showed increased energy efficiency compared to single-session models trained from scratch, consuming $5.04 \mu J$ and $5.03 \mu J$ per inference for the CO and RT tasks, respectively 
(see Fig. \ref{fig:ss_vs_msf}B). \emph{This is equivalent to saving 1.91\% and 2.01\% 
energy per inference for the CO and RT tasks, respectively.} 
\revisions{Together, these results suggest that leveraging cross-session representations not only improves decoding accuracy but also enhances energy efficiency compared to training single-session models from scratch.}

\subsection{Transferring \spikachump to new subjects} \label{sec:transf_to_new_subj}

Having established that pretraining Spikachu on large amounts of data enhances decoding performance and energy efficiency,
we sought to determine whether the pretrained model could be effectively transferred to entirely new sessions from a previously unseen subject. To test this, we trained single-session models from scratch on 12 held-out recording sessions (6 CO and 6 RT) from a new animal (monkey T) drawn from  \citet{Perich2018}. These models achieved a mean test set $R^{2}$ of 0.76 and 0.66 for the CO and RT tasks, while consuming $4.97 \mu J$ and $5.14 \mu J$ per inference, respectively. 

\begin{figure}[h]
    \centering
    \includegraphics[scale=0.43]{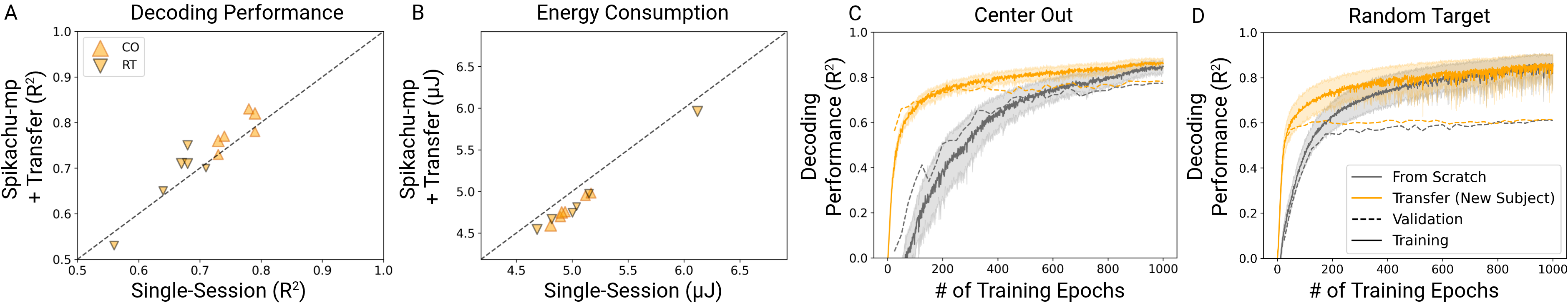} 
    \caption{\textit{Performance comparison between single-session models trained from scratch vs single-session model transferred from \spikachump.} \textbf{(A)}. Decoding Performance, \textbf{(B)} Energy consumption per inference, \textbf{(C, D)} Learning dynamics for the CO and RT tasks.}
    \label{fig:transferring_to_new_subjects}
\end{figure}

We then used \spikachump to transfer the learned representations to these new sessions. The transferred models yielded improved decoding performance, with average per-session test set $R^{2}$ of 0.78 and 0.68 for the CO and RT tasks, respectively (Fig. \ref{fig:transferring_to_new_subjects}A).
In addition to higher decoding accuracy, the transferred models were also more energy-efficient, consuming $4.79 \mu J$ and $4.95 \mu J$ per inference (Fig. \ref{fig:transferring_to_new_subjects}B).
\textit{This represents a reduction in energy consumption of 3.71\% and 3.63\% for the CO and RT tasks, respectively, relative to models trained from scratch}. Finally, we analyzed the training dynamics and observed that the transferred models converged (reached 90\% of their maximum attained $R^{2}$) on average 3× and 4× faster than their scratch-trained counterparts for the CO and RT tasks, respectively (see Fig. \ref{fig:transferring_to_new_subjects}C, D).
These findings suggest that \spikachump learned neural representations that generalize to new subjects, providing a powerful foundation for transfer learning in BCI. 
In practical terms, this means that new users could benefit from robust BCI performance with minimal calibration, reducing the burden of subject-specific training and enabling faster deployment in real-world clinical or assistive settings.

\subsection{Scaling laws of multi-session, multi-subject training}
\label{sec:scaling_laws}

Having shown that pretraining Spikachu learns robust neural representations that generalize across sessions and subjects, we next investigated its scaling behavior, specifically, how (1) decoding performance, and (2) energy efficiency
change as a function of the amount of data used for pretraining. 
In addition to \spikachump, we trained three multi-session, multi-subject models using 20, 49, and 75 sessions drawn from animals C, J, and M in \citet{Perich2018},
the same subjects used to train \spikachump.
To probe generalization and transfer capabilities, we used each pretrained model as the initialization for three finetuning conditions: (1) Seen sessions: finetuning on the same sessions used during pretraining, (2) New sessions: transferring to unseen sessions from the same subjects (NHPs C, J, M), (3) New subject: transferring to unseen sessions from a new subject (monkey T). 

\begin{figure}[h]
    \centering
    \includegraphics[scale=0.43]{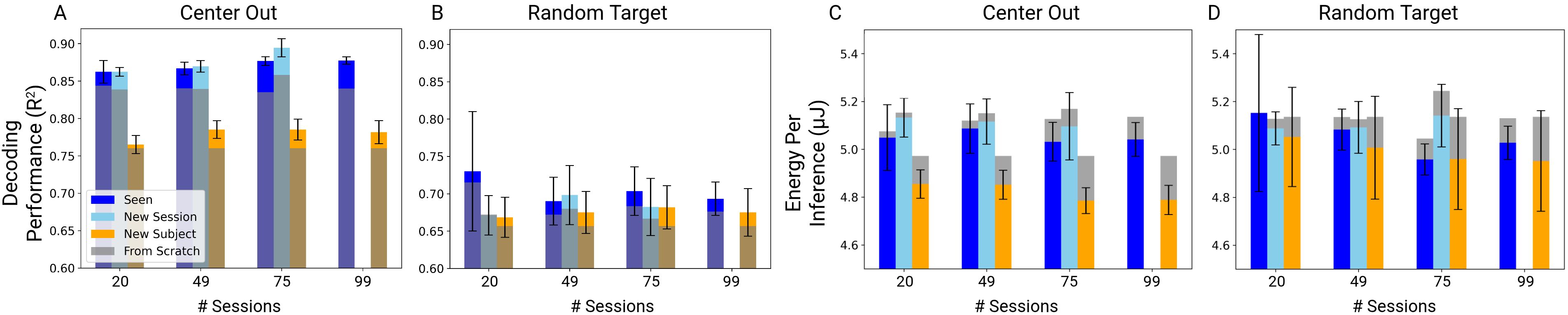} 
    \caption{\textit{Benefits of scaling up the training dataset size.} \textbf{(A, B)} Decoding performance of finetuned and transferred models (performance of scratch-trained single-session models overlayed in gray) as a function of the number of sessions used for model pretraining for the (A) CO, and (B) RT tasks. Panels \textbf{(C, D)} show the energy consumption per inference for finetuned and transferred models (performance of scratch-trained single-session models overlayed in gray) as a function of the number of sessions used for model pretraining for the (C) CO, and (D) RT tasks.}
    \label{fig:benefits_of_scale}
\end{figure}

Decoding performance and energy consumption results for the CO and RT tasks are shown in Fig. \ref{fig:benefits_of_scale}A, B and Fig. \ref{fig:benefits_of_scale}C, D, respectively. For comparison, we overlayed the performance of single-session models trained from scratch (gray bars). Across all conditions, seen sessions, new sessions, and new subjects, the pretrained models consistently outperformed their from-scratch counterparts. Moreover, \emph{the performance gains from pretraining scaled positively with the number of sessions used during model pretraining} (as seen by the growing gap between colored and gray bars in Fig. \ref{fig:benefits_of_scale}A, B, and Fig.~\ref{fig:performance_difference_when_scaling_up}A, B). 
We observed a similar trend for energy consumption per inference, reported in Fig.~\ref{fig:benefits_of_scale}C, D. Pretrained models required less energy across all transfer settings when compared to training from scratch, and \emph{energy savings improved with the size of the pretraining dataset} (Fig.~\ref{fig:performance_difference_when_scaling_up}C, D). As a bonus, \emph{we observed that all pretrained models converged much faster than models trained from scratch (see App. \ref{sec:scaling_laws_of_multi_subject_training_continued})}. 
\revisions{Overall, these results show that scaling up pretraining yielded consistent improvements in both decoding accuracy and energy efficiency, with larger datasets providing greater gains and faster convergence.}

For direct, head-to-head comparisons of decoding performance, energy efficiency, and convergence speed between pretrained and from-scratch models, refer to App. \ref{sec:scaling_laws_of_multi_subject_training_continued} (Fig.~\ref{fig:scaling_analysis_performance} and ~\ref{fig:scaling_analysis_learning_curves}).

\subsection{Transferring \spikachump to a new animal, setup, and task} \label{sec:mc-rtt}

\begin{wrapfigure}{r}{0.5\textwidth} 
    \centering
    \includegraphics[scale=0.5]{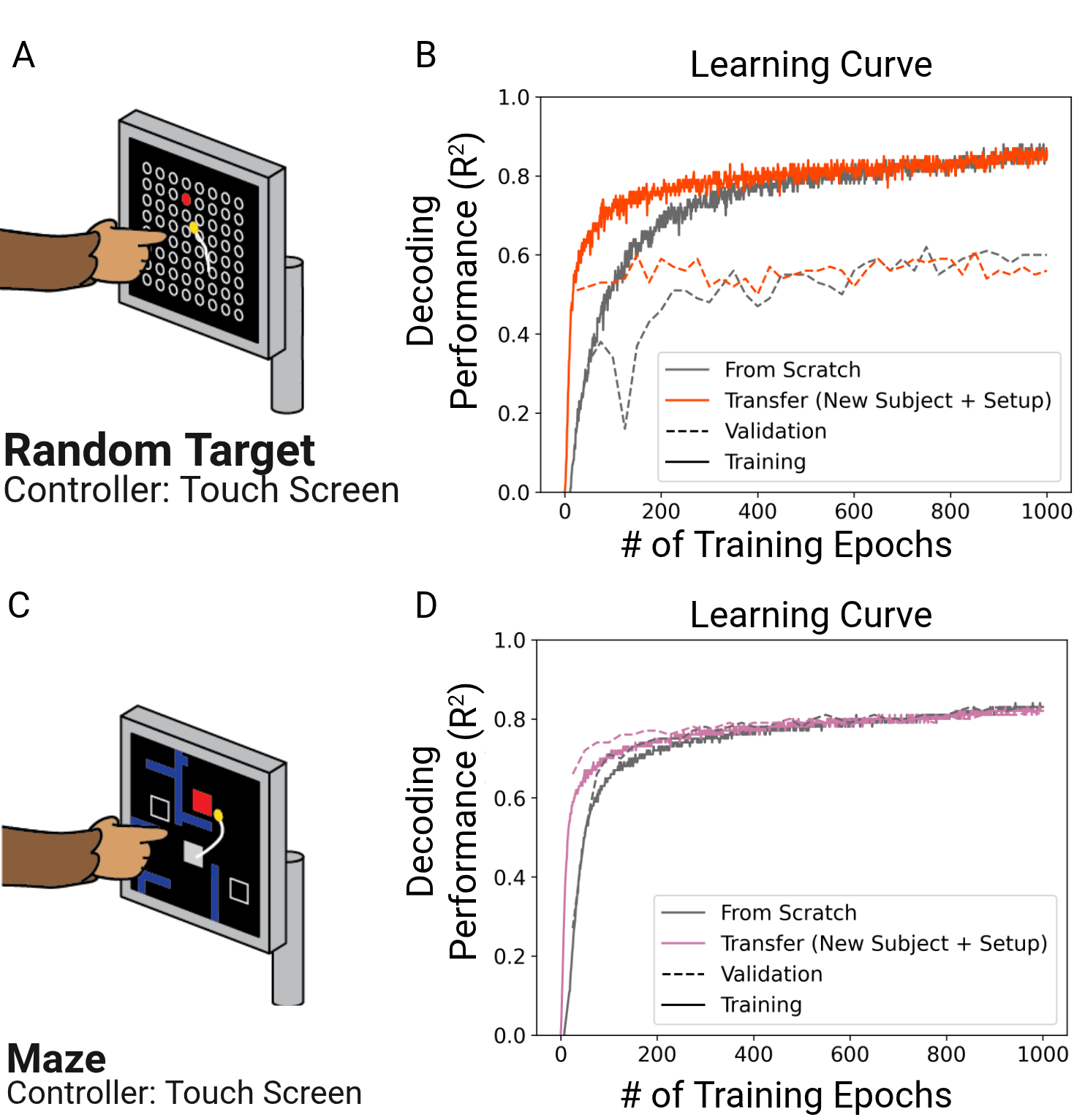}
    \caption{\textit{Generalizing to new animals, setup, and tasks} \textbf{(A, C)} Schematic (reproduced from \citet{Azabou2023}, with permission) and \textbf{(B, D)} learning dynamics for the MC-RTT  and MC-Maze tasks, respectively.}
    \label{fig:nlb_figure}
\end{wrapfigure}

\revisions{In Sec. \ref{sec:transf_to_new_subj} and \ref{sec:scaling_laws}, we demonstrated Spikachu’s ability to generalize across sessions and animals not encountered during pretraining.} 
\revisions{To further assess Spikachu’s generalization capabilities, we evaluated the model on entirely novel conditions: neural recordings from a new animal performing a novel task under experimental settings that substantially differed from those described in \citet{Perich2018}. Specifically, we applied
our framework to the held-out MC-RTT and MC-Maze datasets from the Neural Latents Benchmark \citep{Pei2021nlb} (see Fig. \ref{fig:nlb_figure}A, C for schematics and App. \ref{appendix:sec:data} for task details).}

\revisions{We first trained single-session models from scratch. We used a batch size of 128 for the MC-RTT dataset and a batch size of 512 for the MC-Maze dataset (as this dataset is much larger). 
Spikachu achieved a test set $R^{2}$ and 0.57 on the RTT task and 0.79 on the Maze task, matching the performance of other state-of-the-art models (see \citet{Azabou2023} for a comparison). Those models consumed 5.12 and 4.86 $\mu J$ per inference for the MC-RTT and MC-Maze tasks, respectively.
We then used the pretrained weights of \spikachump as a basis for transferring to the datasets. 
The transferred model achieved $R^{2}$ values of 0.56 for RTT and 0.78 for the Maze task while consuming 4.96 and 4.75 $\mu J$ per inference, respectively.}

\revisions{While this approach did not yield a significant performance gain, \textit{it led to markedly faster convergence (2.33× speedup for the MC-RTT and 2.66× for MC-Maze; see Fig. \ref{fig:nlb_figure}B, D) and reduced energy consumption by 3.03\% and 2.25\% for the MC-RTT and MC-Maze tasks, respectively}. This result is particularly noteworthy given the substantial difference in experimental conditions between the MC-RTT and MC-Maze datasets and those used in pretraining \spikachump. Despite these domain shifts, the neural representations learned through pretraining proved transferable, highlighting Spikachu’s ability to generalize across animals, recording setups, and tasks.}

\section{Discussion}
\label{sec:discussion}

In this work, we introduced Spikachu: a causal, scalable, and energy-efficient framework for multi-session, multi-subject neural decoding based on SNNs. Unlike existing approaches, it combines the expressivity of transformers with the low-power advantages of SNNs, resulting in a system that is both high-performing and energy-conscious.
Our experiments demonstrate that Spikachu performs comparably to state-of-the-art neural decoding models while offering large energy savings. 
\revisions{We also demonstrate that pretraining our model on a large corpus of heterogenous neural data benefits neural decoding.} 
To our knowledge, this is the first demonstration that unified models trained on multi-subject, multi-task neural data can yield improved performance in the spiking domain. Moreover, we find that pretraining encourages sparsity in network activity, further enhancing energy efficiency, a critical factor for scaling up training across heterogeneous datasets. %effective neural decoders designed for implantable BCIs. %We also show that pretrained models can be transferred to new sessions and animals very efficiently.

This work provides strong experimental evidence that SNNs are a practical and energy-efficient alternative to ANNs for neural decoding. While the majority of our architecture leverages SNNs to achieve substantial energy savings, it still relies on an ANN-based harmonization module \revisions{to support large-scale training across datasets and subjects.}
Although this module represents a small fraction of the total model parameters \revisions{(less than 15\% of total parameters)}, it accounts for a disproportionately high share of the model’s energy.
Replacing it with a fully spiking alternative could yield additional energy savings. We believe that adaptations of spiking attention, such as those introduced by \citet{zhou2024spikformerV2} or \citet{Li2024}, offer promising pathways to develop a fully spiking, end-to-end architecture.

Although our results provide strong evidence for spiking-based neural decoding, real-world validation remains a critical next step. Implementing Spikachu on neuromorphic hardware and using it in online experiments would ultimately test its usability in the real world. Prior works have demonstrated the feasibility of simple online SNN decoders \cite{Dethier2011, Liao2022, Leone2023, Taeckens2023} \revisions{but scaling to more complex architectures remains a formidable challenge. Encouragingly, Spikachu’s resource requirements in this work are modest (fewer than 4M synapses and roughly 10K neurons) and well within the capacity of modern neuromorphic chips such as the Loihi 2 \citep{orchard2021} and Darwin 3 \citep{Ma2024Darwin}. Importantly, the compact form factor of these chips is also well within the physical constraints of fully implantable BCI systems, which typically fit within the area of a U.S. quarter.}
We hope our work inspires collaboration between neuroscience and hardware communities to address implementation barriers.

Overall, this work represents a meaningful step forward in neuroscience by introducing an energy efficient neural decoding framework based on SNNs.%, all essential for real-world neural decoders that run on battery-constrained, implantable BCIs. %Unlike most existing approaches, we combine the representational power and scalability of transformers with the low-power advantages of spiking neurons, resulting in a system that is both high-performing and energy-conscious.
We demonstrate that our framework delivers strong decoding performance that scales when trained on large, heterogeneous, multi-session datasets. This challenges the conventional notion that spiking networks must sacrifice performance for energy savings and highlights their potential for broader generalization \citep{zhou2024spikformerV2}. Given that early demonstrations of BCI therapeutic interventions have already been shown \citep{Metzger2023, Collinger2013, Flesher2021, Willsey2025, Willett2023}, our work paves the way towards energy efficient neural decoding therapeutic interventions, bringing them closer to clinical translation.

\begin{ack}

We would like to thank the following funding sources for their generous support: National Institutes of Health grant R01NS121219 (FV);  National Science Foundation CAREER award IIS-214607 (ELD);
Onassis Foundation graduate student scholarship and A. G. Leventis Foundation graduate student scholarship (GM);
Bodossaki Foundation graduate student scholarship (IA).

Authors declare no competing interests.
\end{ack}

\bibliographystyle{unsrtnat}
\bibliography{actual_citations}  % your .bib file, no .bib extension

\clearpage

\section*{Appendix}

\appendix

\section{Concept Primers} \label{app:concept_primers}

\subsection{Spiking neuron models}
\label{app:snns}

\paragraph{Overview of spiking model history.}
The simplest neuron model available for machine learning is the standard ANN neuron, which statelessly aggregates incoming signals with addition and then passes the result through a predetermined non-linearity \citep{non-lin-analysis}.
Bio-inspired models retain more of the complexity of biological neurons at the cost of being more technically involved.
Amongst the first to quantify such models, \citet{hodgkin-huxley-squid} developed a multi-parameter model to describe squid axon neurons.
Other works have provided detailed models to approximate different types of biological neurons.
For instance, \citet{migliore-pyramidal-bp} and \citet{herrera-pyramidal} modeled various aspects of pyramidal neurons, whereas \citet{mulloney-motor-neurons} modeled motor neurons in the swimmeret system of crayfish.
\footnote{A collection of computational neuroscience models can be retrieved at: \url{https://modeldb.science/}.}

\paragraph{The Leaky Integrate-and-Fire neuron.}
Between the simplicity of ANN neurons and fully biological neurons lie several models that trade off between the two, retaining the statefulness of biological systems while greatly simplifying the computational specifics to a much more tractable form.
These models, generally termed \textit{spiking neurons}, include state variables that typically correspond to potentials of various ion channels in biological systems.
They also contain logic that dictates the evolution of these variables over time, typically expressed as a differential equation, as well as the necessary conditions for firing output spikes.
The most ubiquitous such model is the Leaky Integrate-and-Fire (LIF) model, which approximates a neuron with a spiking RC circuit.
In this model, the tracked state is a real, scalar number corresponding to the membrane voltage of the neuron.
The state $v(t)$ obeys its own recurrent dynamics, modeled as an exponential decay, to which inputs are added through integration.
The differential equation for the potential with time-varying input current $i(t)$,
\begin{gather}
    \tau \cdot \dot v(t) = -v(t) + R \cdot i(t), ~
    \label{eqn:lif-membrane}
\end{gather}
can be solved analytically,
\begin{gather}
    v(t) = e^{-t / \tau} \cdot \left(
        v_0 + \int_{t_0}^t e^{s/\tau} \cdot \frac{R \cdot i(s)}{\tau} ds
    \right),
\end{gather}
where the parameter $\tau$ controls the decay rate and $R$ corresponds to an input resistance.
A LIF neuron's firing behavior is then a simple matter of thresholding, %as shown in Eq. \ref{eqn:lif-fire}.
\begin{gather}
    s(t) = \mathds{1}[v(t) > V_{th}] , ~
    v(t + \delta t) = \begin{cases}
        v(t) + \dot v(t) \cdot \delta t &\textrm{, if } s(t) = 0 \\
        V_{rs} &\textrm{, if } s(t) = 1
    \end{cases},
    \label{eqn:lif-fire}
\end{gather}
where $s(t)$ is the spike output and $\mathds{1}$ is the indicator function.
If the membrane voltage is above the threshold $V_{th}$, there will be a binary spike in the output, and the state will be reset according to the reset logic; one way to implement this would be to set the membrane back to a resting potential value $V_{rs}$, typically zero.
Recently, it has been shown that these parameters can be \emph{learned} \citep{fang2020learnable-membranes}.
We utilize this property in the present work to specialize different network components for separate (implicit) portions of the input signal.

\paragraph{Other spiking neurons.}
Other neuron models have found application in SNNs.
In \citet{orchard2021}, Resonate-and-Fire (RF) spiking neurons were used.
These neurons are very similar to LIF, except that their state is \textit{complex}, meaning that excitation induces an oscillatory behavior.
Spiking only occurs when the imaginary part of the membrane state is zero, and the output is not binary, but instead the instantaneous value of the real part of the state.
It is eventually shown that the resulting behavior is mathematically similar to the Short-Time Fourier Transform (STFT).
Similarly, \citet{xing2025spikellm} introduced a generalized version of the LIF neuron. With integral outputs, the Generalized Integrate-and-Fire (GIF)  neuron supported the authors' saliency-based spiking large language models.
These works showcase the different directions in which spiking neurons can develop.

\subsection{Spiking vs vanilla self attention} \label{sec:vsa_vs_ssa}

In this work, we use spiking self-attention (SSA) introduced by \citet{zhou2023spikformer} (an extension of the vanilla self-attention (VSA) introduced by \citet{Vaswani2017}) which was specifically designed for SNNs. 
Here, we provide a brief overview of the differences between the two mechanisms.

\paragraph{Vanilla self attention.}

Let an input sequence $X \in \mathbb{R}^{T \times N \times D}$, where $T, ~N, ~D$ refer to the number of time steps, the number of tokens in the sequence, and the dimensionality of each token, respectively. VSA relies on three components,
\begin{equation}
    \text{Queries: }\mathbf{Q} = \mathbf{W}_{Q} \mathbf{X}, ~~~~ \text{Keys: }\mathbf{K} = \mathbf{W}_{K} \mathbf{X}, ~~~~ \text{and Values: } \mathbf{V} = \mathbf{W}_{V} \mathbf{X},
\end{equation}
and is computed as,
\begin{equation}
    \text{VSA}(\mathbf{Q}, \mathbf{K}, \mathbf{V}) = \text{Softmax}\left(
        \frac{\mathbf{QK}^{\top}}{\sqrt{d_{k}}}
    \right) \mathbf{V},
    \label{eq:attn}
\end{equation}
where $d_{k}$ refers to the feature dimensionality of $\mathbf{K}$. %The operation is typically For this operation, we use the standard transformer block precluded by layer-normalization and followed by a feed-forward network \citep{Vaswani2017}. 

VSA can be directly used on spiking sequences. However, it is unsuitable because: (1) the float-point
matrix multiplication of $\mathbf{Q}$, $\mathbf{K}$ and softmax function (which contains exponent calculation and
division operation) do not comply with the calculation rules of SNNs, and (2) computational and space requirements for VSA scale quadratically with the length of the input sequence $\mathbf{X}$ which does not meet the efficient computational requirements of SNNs \cite{zhou2023spikformer}.

\paragraph{Spiking self attention.}
\label{app:ssa-definition}

\citet{zhou2023spikformer} proposed SSA, an alternative to VSA which is more suitable for SNNs. In their formulation, the sequence $\mathbf{X} \in \mathbb{R}^{T \times N \times D}$ is projected into equally shaped Queries ($\mathbf{Q}$), Keys ($\mathbf{K}$), and Values ($\mathbf{V}$),

\begin{equation}
    \mathbf{Q} = \mathcal{SN}_{Q} (\textrm{BN} (\mathbf{W}_{Q}\mathbf{X})), ~~~~ \mathbf{K} = \mathcal{SN}_{K} (\textrm{BN} (\mathbf{W}_{K} \mathbf{X})), ~~~~ \mathbf{V} = \mathcal{SN}_{V} (\textrm{BN} (\mathbf{W}_{V}\mathbf{X})),
\end{equation}
and compute the spiking self-attention matrix as,
\begin{equation}
    \text{SSA}^\prime = \mathcal{SN}(\mathbf{QK}^\top \mathbf{V} \cdot s),
    \label{eq:ssa}
\end{equation}
where $s$ is a scaling factor.
SSA is then computed via the following spiking layers,
\begin{equation}
     \textrm{SSA}(\mathbf{Q}, \mathbf{K}, \mathbf{V}) = \mathcal{SN}(\textrm{BN}(\mathbf{W}_{\textrm{ssa}}~ \textrm{SSA}^{\prime})).
\end{equation}

This formulation is better suited for spiking sequences than traditional VSA for several reasons. First, SSA operates independently at each time step, aligning naturally with the temporal dynamics of SNNs. Unlike VSA, SSA removes the use of the softmax function to normalize the attention matrix. Instead, it directly computes the attention by multiplying the query, key, and value sequences 
without additional normalization.
This simplification is more efficient as softmax introduces unnecessary computational complexity. Since the spiking neuron layers $\mathcal{SN}_{Q}$ and $\mathcal{SN}_{K}$ output binary spike trains, the resulting attention maps are inherently non-negative. This removes the need for softmax to enforce non-negativity, as the attention mechanism already emphasizes relevant features and suppresses irrelevant ones through sparse, event-driven computation.

\newpage

\section{Datasets}
\label{appendix:sec:data}

To train and evaluate our framework, we leveraged a diverse collection of publicly available datasets spanning multiple NHPs and various behavioral paradigms.

\subsection{Datasets used for training and validation}

To develop our model, we leveraged a rich and diverse dataset comprising 99 unique recording sessions from three NHPs (monkeys C, J, and M) engaged in two distinct behavioral tasks. These recordings were collected across four foundational studies \cite{perich2017altered, glaser2018population, Perich2018, gallego2020long}, and were later curated into a unified dataset by \citet{Azabou2023}. This comprehensive resource is publicly available through \href{https://dandiarchive.org/}{Dandi} (used in this work) and via the \href{https://brainsets.readthedocs.io/en/latest/index.html#}{\texttt{brainsets}} platform.

Across all sessions in the dataset, each NHP was seated in a primate chair, and a custom two-dimensional planar manipulandum was used to control a cursor displayed on a computer screen. During each recording session, the NHP performed one of two structured motor tasks:

\begin{itemize}
    \item \textit{Center-out Task (CO):} In this classic reaching task, the monkey initiated movement from a central target toward one of eight peripheral targets arranged uniformly around a circle with an 8~cm radius. After holding at the center target for a variable duration, an auditory ``go'' cue signaled the monkey to move to a designated outer target. This task is widely used to investigate neural processes underlying movement planning, preparation, and execution.
    
    \item \textit{Random Target Task (RT):} Similar in design to the CO task, the RT task involved targets that were randomly distributed across the workspace rather than arranged in a circular pattern. The monkey was instructed, via an auditory cue, to move sequentially between four randomly positioned targets, introducing greater variability in movement direction and distance.
\end{itemize}

The behavioral sampling rate used across all sessions within this dataset was 100 Hz. We used 70\% of the data from within each session for training and 10\% for validation; the remaining 20\% was held-out for testing.

\subsection{Datasets held out for testing}

We reserved 20\% of the data from each session described above for testing. In addition, we fully held out all sessions from Monkey T, comprising 6 CO and 6 RT sessions, to assess the model’s ability to generalize across subjects. 

To asses the model's ability to generalize to novel tasks, we also held-out two standardized datasets from the Neural Latents Benchmark~\cite{Pei2021nlb}, MC-Maze and MC-RTT, which are described in detail below. In both datasets, the animals (different for MC-Maze and MC-RTT) performed the tasks by swiping on a touch screen instead of using a manipulandum.

\begin{itemize}
    \item \textit{MC Maze:} In this task, the monkey performed delayed reaches to visually presented targets while navigating around the boundaries of a virtual maze. The dataset includes a wide range of behavioral configurations, each defined by unique combinations of target positions, barrier counts, and barrier placements. This diversity results in both straight and curved reach trajectories. With thousands of trials, MC Maze provides a rich substrate for analyzing population-level neural dynamics. Moreover, its delayed reaching design facilitates a clear dissociation between neural activity related to movement preparation and execution.
    
    \item \textit{Random Target Task (RTT):} In this task, the monkey executes a variant of the RT task described previously. Instead of performing reaches separated via auditory cues, the animal completes continuous, point-to-point reaching movements between virtually presented targets. The movements begin and end at various locations, span variable distances, and include only a few repeated trajectories. This variability makes RTT particularly useful for evaluating a model’s ability to generalize across complex and less stereotyped behaviors.
\end{itemize}

The behavioral sampling rate for the RTT and Maze tasks was 1000 and 100 Hz, respectively. To standardize the sampling rate across all datasets used in this work, we downsampled behavioral samples for the RTT task to 100 Hz.

\newpage

\section {Model Implementation Details} \label{app:model_implementation_details}

\subsection{Spikachu implementation details}
\label{sec:spikachu_details}

Spikachu is a general framework that supports various implementations. Here, we outline the one used in this work.

\paragraph{Learnable embeddings for units.}
We assign each unit a unique 32-dimensional learnable embedding.

\paragraph{Harmonizer.}
This is the ANN part of our architecture described in Sec. \ref{sec:harmonizer}. The cross-attention block is implemented as described in \citet{Jiagle2021Perceiver} and is initialized with 128, 32-dimensional learnable latent queries. The linear layer following the cross-attention block projects the latents to a 128-dimensional vector.  

\paragraph{Multi Scale SNN-I.} The 128-dimensional latent vector is passed through 3 parallel spiking MLPs. Each MLP is composed of 4 layers and projects the 128-dimensional input vector to a 256-dimensional latent vector (all hidden layers and output layer have a dimensionality of 256). Each MLP has neurons whose membrane potential dynamics evolve at distinct temporal scales. We initialize the learnable decay constants of the neurons of the 3 parallel networks to values $\tau$ of 1.11, 1.46, and 434.79 and allow them to update freely during model training.

\paragraph{Spiking Self-Attention.}
The three 256-dimensional latents extracted by Multi Scale SNN-I are arranged in a sequence and processed using spiking self-attention as described by \citet{zhou2023spikformer}. The input is projected to equally shaped 512-dimensional keys, queries, and values prior to spiking self-attention which is carried out using 8 heads. The latent is then passed through a 2-layer spiking feed forward network with a hidden size of 256. Residual connections are employed after the attention calculation as well as after the feed forward network, as is employed in Spikformer \citep{zhou2023spikformer}.  

\paragraph{Spiking MLP.} The three 256-dimensional outputs of the attention operation are unrolled and are projected to a dimensionality of 384 via one linear layer followed by a spiking activation. 

\paragraph{Multi Scale SNN-II.} The 384-dimensional latent is passed through 2 parallel spiking MLPs composed of 4 layers each. The hidden size for each MLP is set to 384. The learnable decay constant for the neurons of each of the two parallel MLPs is set to an initial $\tau$ of 1.11, and 434.79. Following processing, the latents from the 2 parallel MLPs are concatenated, resulting in a latent with dimensionality 768.

\paragraph{Membrane potential observer layer.} The 768-dimensional latent is then passed through a spiking activation. Instead of tracking the output of the neurons of this activation layer (they never fire), we keep track of their membrane potential instead. This results in a latent of the same dimensionality of 768.

\paragraph{Readout layer.} The 768-dimensional latent is then projected to a 2-dimensional vector, which is our network's velocity prediction ($v_{x}$, $v_{y}$), using a linear layer.

\subsection{Baseline model implementation details}
\label{app:baselines}

We introduced several baseline architectures for our experimental comparisons, which we describe in detail below.

\paragraph{MLP.}
Our MLP baseline comprises seven consecutive hidden dense layers of ANN neurons, with input shapes: $[100,256,512,1024,1024,512,256]$.
We used the ReLU nonlinearity and dropout probability of $0.1$.
Since the model is not recurrent, we supplied a rolling window of 10 consecutive binned spike windows during inference.

\paragraph{GRU.}
Our GRU baseline comprises four sequential blocks, as implemented in \texttt{PyTorch}.
Each block contains four cells, with a hidden dimension of 164.
We used a dropout probability of $0.1$.

\paragraph{LSTM.}
Our LSTM baseline comprises four sequential blocks, as implemented in \texttt{PyTorch}.
Each block contains four cells, with a hidden dimension of 162.
We used a dropout probability of $0.1$.

\paragraph{POYO.}
\label{app:poyo}
Our POYO baseline was implemented using the model coded by \citet{Azabou2023}, made available at \href{https://github.com/neuro-galaxy/poyo}{https://github.com/neuro-galaxy/poyo}. We trained the model using N=256 latent tokens each with a dimensionality of 128. 

\begin{figure}[h]
    \centering
    \includegraphics[scale=0.5]{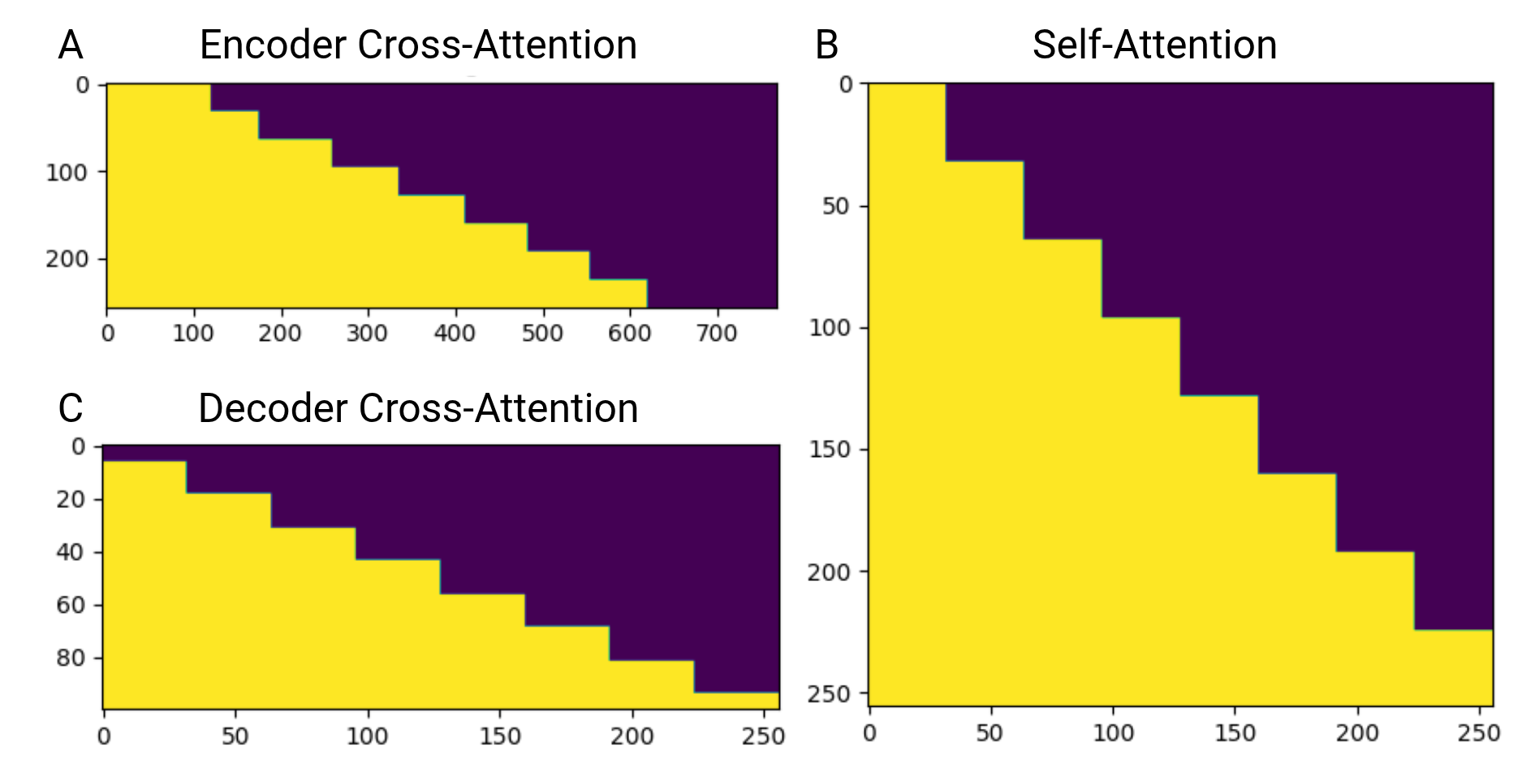} 
    \caption{\textit{Masking used to make POYO Causal.} Mask used in \textbf{(A)} encoder cross-attention, \textbf{(B)} self-attention, and \textbf{(C)} decoder cross-attention blocks.}
    \label{fig:causal-poyo-masks}
\end{figure}

\paragraph{Causal POYO.}
Our causal POYO baseline is implemented using the model coded by \citet{Azabou2023}. On top of the provided model, we added causal masking in all the model's transformer blocks. Specifically, for each attention module described in \citet{Azabou2023}, let $\mathbf{Q}$ be the set of $N$ latent query vectors with timestamps $t^Q_i, i=1,\dots,N$, and $\mathbf{K}$ be the set of $M$ key vectors with timestamps $t^K_i, i=1,\dots,M$. We dynamically masked the attention multiplication $\mathbf{Q} \mathbf{K}^\top$ such that $t^Q_i \leq t^K_j$, for all pairs $i,j$ for which the attention matrix is computed.
This is a direct generalization of lower-triangular masking of the attention matrix commonly used in transformers to prevent tokens from accessing future tokens, effectively making the operation causal. 
The masks used for masking the (1) encoder cross-attention block, (2) each self-attention block, and (3) the decoder cross-attention block used in causal POYO are shown in  Fig. \ref{fig:causal-poyo-masks}.

\newpage

\section{Model Training Details}
\label{app:spikachu-training}

All models were coded in PyTorch \citep{paszke2019pytorch}. The spiking components of Spikachu were implemented using SpikingJelly \cite{fang2023spikingJelly}.

\subsection{Training hyperparameters}

All models were trained using LAMB optimizer \citep{You2019lamb} with weight decay set to $10^{-4}$. The learning rate was initialized at $2 \times 10^{-3}$ and held constant for the first 75\% of training epochs and was decayed to zero using a cosine schedule for the remaining 25\%. Single-session models were trained for 1000 epochs with a batch size of 128, while multi-session models were trained for 400 epochs with a batch size of 512.

\subsection{Compute}
\label{app:compute}

Model training was conducted on a multi-GPU cluster. Depending on resource availability, single-session models were trained using one of several NVIDIA GPUs, from the following: L40s, L40, A40, 3090, A6000, 2080 Ti, and A10. Training these models required under 11GB of GPU memory and typically less than an hour, with a few sessions taking up to 5 hours when trained on a 2080 Ti graphics card. Finetuning to individual sessions was performed on the same range of GPUs and took approximately the same amount of time as training from scratch.

Training of multi-session models was performed on an NVIDIA A40 GPU with 48GB of memory. All multi-session models, including \spikachump, our most computationally intensive model, completed training in under 48 hours.

\subsection{Data augmentation}

For all model training (both single-session and multi-session) we employed unit dropout augmentation, introduced by \citet{Azabou2023}. In this approach, a random subset of the recorded neural population is sampled within time windows and used for training, effectively simulating variability in unit availability. To prevent the augmentation from being overly destructive, we enforced a minimum population size of 30 units per window. %Importantly, this threshold applies only during training; the model itself remains capable of ingesting arbitrarily large populations of neural units.

\subsection{Training objective}

All models were trained by minimizing the MSE loss between the true and predicted hand velocity sequences of the NHPs performing the various behavioral tasks. For the RT task, all behavioral timepoints were weighted equally during training. For the CO task, we increased the weight of the prediction during the reaching segments of the movements by a factor of 5 as has been done in previous works \citep{Azabou2023, brown2020language}. 

\subsection{Evaluation details}

Although model training is performed on arbitrary segments of neural data, evaluation follows standardized protocols established in prior works \citep{Azabou2023, Pei2021nlb, pandarinath2018latent}. For CO and MC-Maze, we report decoding performance only during the reaching phase of successfully completed trials. For RT, which includes a hold period followed by a sequence of 3 to 4 random reaches, we likewise evaluate performance only during the reaching movements. Since movements are continuous for MC-RTT, we report decoding accuracy across all available segments.

\subsection{Surrogate gradients}

Spiking neurons are inherently non-differentiable due to their discrete activation function (the Heaviside step function) which presents a fundamental challenge for gradient-based backpropagation. To overcome this limitation, we adopt a surrogate gradient approach, replacing the non-differentiable Heaviside function with the arctangent (\texttt{atan}) function when calculating the gradients used for backpropagation. This surrogate function has been demonstrated to support the stable training of deep SNNs in prior works \cite{fang2023spikingJelly, eshraghian2023snntorch}.

\subsection{Smoothing}

\revisions{Prior to gradient computation, model predictions were smoothed along the temporal axis using a moving average with a window size of 20 samples. To ensure a fair comparison, identical smoothing was used across all models.}

\subsection{Finetuning/Transferring details}

The training procedure for finetuning and transferring \spikachump to individual sessions is identical to that of training single-session models from scratch, except: models trained from scratch were initialized randomly (using seed 42); finetuned and transferred models were initialized with the pretrained weights from the multi-session pretrained models. In both cases, all learnable parameters were updated throughout training.

\newpage

\section{Energy Analysis}
\label{app:energy}

Estimating energy consumption for any machine learning model consists of approximating the total amount of floating-point operations (FLOPs) necessary to execute the forward pass of the model.
These are split into two categories, multiply-accumulate (MAC) and accumulate (AC) operations.
A MAC operation multiplies two operands and adds the result to an accumulator.
Symbolically, if $x, y$ are the operands and $a$ is the accumulator, a MAC is defined as: $a \leftarrow a + x \cdot y$.
An AC operation acts on a single operand by adding it to an accumulator.
If $x$ is the operand and $a$ is the accumulator, an AC is defined as: $a \leftarrow a + x$.

Each instance of these operations consumes a certain amount of energy, and by summing, we can estimate the total energy consumption of a model's forward pass.
As quantified by \citet{horowitz2014}, performing ACs is cheaper than MACs in hardware, specifically: $E_{\textrm{MAC}} \simeq 4.6pJ$, $E_{\textrm{AC}} \simeq 0.9 pJ$.
Therefore, a model's total energy becomes: $E_{\textrm{total}} = N_\textrm{AC} \cdot E_\textrm{AC} + N_\textrm{MAC} \cdot E_\textrm{MAC}$.
The $N$ variables represent the total count of the respective operations in the architecture.

For reference, a matrix multiplication of the form $A \cdot B,~ A \in \mathbb{R}^{n \times m}, B\in \mathbb{R}^{m \times k}$, requires $(n \cdot m \cdot k)$ MAC operations, and vector addition of dimension $d$ requires $d$ AC operations.

\subsection{Baseline energy calculation}
\label{app:energy-baselines}

Here, we describe the methodology used to compute the energy required per inference for each baseline model in this work.

\subsubsection{MLP}
\label{app:baselines-mlp}
Each linear layer of an MLP is parameterized by a weight matrix $\mathbf{W} \in \mathbb{R}^{n_i \times n_o}$ and a bias vector $\mathbf{b} \in \mathbb{R}^{n_o}$.
Given an input $\mathbf{x} \in \mathbb{R}^{n_i}$, the operation performed is: $\mathbf{z = Wx + b}$.
The energy expenditure of an $L$-layer MLP is therefore,
\begin{gather}
    E_{\textrm{MLP}} = \sum_{l=1}^{L} 
    n^{(l)}_i n^{(l)}_o \cdot E_{\textrm{MAC}} + n^{(l)}_o \cdot E_{\textrm{AC}},
    \label{eq:mlp-energy}
\end{gather}
where $l \in \{1, \dots, L\}$ is the layer index.

\subsubsection{GRU}
The equations for a GRU cell are described in detail by \citet{chung2014gru}. Briefly, each cell can be expressed as,
\begin{align*}
    r_t &= \sigma(W_{ir} x_t + b_{ir} + W_{hr} h_{(t-1)} + b_{hr}), \\
    z_t &= \sigma(W_{iz} x_t + b_{iz} + W_{hz} h_{(t-1)} + b_{hz}), \\
    n_t &= \tanh(W_{in} x_t + b_{in} + r_t \odot (W_{hn} h_{(t-1)} + b_{hn})), \\
    h_t &= (1 - z_t) \odot n_t + z_t \odot h_{(t-1)},
\end{align*}
where $x$ is the input, $W$-variables refer to weight matrices, $b$-variables to bias vectors, $h$ is the hidden state, $t$ the timestep index, $i$ denotes variables acting directly on the input, and $r,z,n$ are the reset, update, and new gates, respectively.
We also denote the sigmoid function with $\sigma$ and the Hadamard product with $\odot$.
From these equations, we can calculate the total operations performed in each cell as a function of its input and hidden state sizes as,
\begin{gather}
    N^\textrm{GRU}_\textrm{MAC} = 3 \cdot (
        n_h (n_i + n_h) + n_h
    ) + 4 n_h, ~\text{and}~ N^{\textrm{GRU}}_\textrm{AC} = 6 n_h,
\end{gather}
where $n_i$ is the input size and $n_h$ is the hidden size.
For the full module comprising multiple cells, we sum the counts of the above operations for each cell to obtain the total energy,
\begin{gather}
    E_\textrm{GRU} = N_B \cdot \left(
        E_\textrm{MAC} \sum_{c=1}^C  N^\textrm{GRU}_\textrm{MAC}(c) + E_\textrm{AC} \sum_{c=1}^C  N^\textrm{GRU}_\textrm{AC}(c)
    \right) + E_\textrm{MLP},
    \label{eq:gru-energy}
\end{gather}
where $c$ is an index over the $C$ cells in each block, $N_B$ is the total number of multi-cell blocks, and $E_\textrm{MLP}$ represents the energy of an MLP used for readout, with energy given by Eq. \ref{eq:mlp-energy}.

\subsubsection{LSTM}
The exact implementations of the LSTM architecture can be found in \citet{sak2014lstm}. Briefly, each cell can be described as,
\begin{align*}
    i_t &= \sigma(W_{ii} x_t + b_{ii} + W_{hi} h_{(t-1)} + b_{hi}), \\
    f_t &= \sigma(W_{if} x_t + b_{if} + W_{hf} h_{(t-1)} + b_{hf}), \\
    g_t &= \tanh(W_{ig} x_t + b_{ig} + W_{hg} h_{(t-1)} + b_{hg}), \\
    o_t &= \sigma(W_{io} x_t + b_{io} + W_{ho} h_{(t-1)} + b_{ho}), \\
    c_t &= f_t \odot c_{(t-1)} + i_t \odot g_t, \\
    h_t &= o_t \odot \tanh(c_t),
\end{align*}
where $x$ is the input, $W$-variables refer to weight matrices, $b$-variables to bias vectors, $h$ is the hidden state, $t$ the timestep index, $i,f,g,o$ are the input, forget, cell, and output gates, respectively, and $c$ is the cell state.
From these equations, the total operations can be counted as,
\begin{gather}
    N^\textrm{LSTM}_\textrm{MAC} = 4 \cdot (
        n_h (n_i + n_h) + n_h
    ) + 3n_h,~  \text{and} ~
    N^\textrm{LSTM}_\textrm{AC} = 9 n_h,
\end{gather}
where $n_i$ is the input size and $n_h$ is the hidden size. For the full module comprising multiple cells, we sum each cell's counts of the above operations to obtain the total energy,
\begin{gather}
    E_\textrm{LSTM} = N_B \cdot \left(
        E_\textrm{MAC} \sum_{c=1}^C  N^\textrm{LSTM}_\textrm{MAC}(c) + E_\textrm{AC} \sum_{c=1}^C  N^\textrm{LSTM}_\textrm{AC}(c)
    \right) + E_\textrm{MLP},
    \label{eq:lstm-energy}
\end{gather}
where $c$ is an index over the $C$ cells in each block, $N_B$ is the total number of multi-cell blocks, and $E_\textrm{MLP}$ represents the energy of a small MLP used for readout, with energy given by Eq. \ref{eq:mlp-energy}.

\subsubsection{POYO}
\label{app:tf-energy}

\paragraph{Attention.}
We will first describe how we calculate the energy for a single attention module, based on which POYO is built \cite{Azabou2023}. 
To compute the energy requirements of the attention operation (see Eq. \ref{eq:attn}), we cache the dimensions of the query, key, and value tensors at runtime.
We can then precisely calculate the FLOPs necessary for the matrix multiplications, scaling, and/or softmax operations performed in attention.
The following derivation explains the energy requirements for a single cross-attention block.
\footnote{For self-attention, we need only set the sequence lengths to be equal; the derivation is otherwise identical.}

Let $L_1, L_2$ be the input and output sequence lengths respectively, $E$ the embedding dimension, $\mathbf{Q} \in \mathbb{R}^{L_1 \times E}$ the query tensor, $\mathbf{K} \in \mathbb{R}^{L_2 \times E}$ the key tensor, and $\mathbf{V} \in \mathbb{R}^{L_2 \times D}$ the value tensor.
The computational requirements following Eq. \ref{eq:attn}, \emph{for a single attention head}, are computed as follows,
\begin{gather*}
    N_{\textrm{MAC}}^{QK^\top} = E L_1 L_2, ~
    N_{\textrm{MAC}}^{\textrm{norm}} = 3 L_1 L_2, ~
    N_{\textrm{MAC}}^{V} = D L_2^2 \Rightarrow \\
    N_{\textrm{MAC}}^{\textrm{CA}} = L_2 \cdot ((E+3)L_1+D L_2).
\end{gather*}
For \emph{multi-head attention}, we need only multiply the above by the number of heads.
Letting $H$ be the number of heads, the total energy for the cross attention operation is,
\begin{gather}
    E_\textrm{CA} = H \cdot E_{\textrm{MAC}} \cdot N_{\textrm{MAC}}^{\textrm{CA}}.
    \label{eq:energy-attn}
\end{gather}
Note that the cost for projecting inputs to $\mathbf{Q}, \mathbf{K}, \mathbf{V}$ and projecting outputs to the appropriate downstream shape is computed separately (linear layers, computation follows the logic of \ref{app:baselines-mlp}).

\paragraph{POYO.} By iteratively applying Eq. \ref{eq:energy-attn}, computing the energy requirements per inference for POYO becomes straightforward.
For each attention block, we need to account for the attention energy, and also the energy necessary for projections, given by Eq. \ref{eq:mlp-energy}.
Specifically,
\begin{gather}
    E_\textrm{proj} = E_\textrm{proj}^Q + E_\textrm{proj}^K + E_\textrm{proj}^V + E_\textrm{proj}^o,
\end{gather}
where the summands correspond to the energy required for the query, key, value, and output projections respectively.
The total energy requirement for POYO can now be expressed as,
\begin{gather}
    E_\textrm{POYO} =
        (
            E_\textrm{CA}^\textrm{enc} +
            E_\textrm{proj}^\textrm{enc}
        ) +
        N_\textrm{SA} (
            E_\textrm{SA} + E_\textrm{proj}^{\textrm{SA}}
        ) +
        (
            E_\textrm{CA}^\textrm{dec} + E_\textrm{proj}^{\textrm{CA}}
        ) + 
        E_{\textrm{MLP}},
\end{gather}
where $N^\textrm{SA}$ is the number of consecutive self-attention blocks used in the model.
We note that one more subcomponent exists for this model, namely positional encoding using RoPE  \citep{Su2024}.
However, since the operations involved in computing the positional encoding are very limited, we did not take them into account when estimating POYO’s energy requirements.

\subsection{SNN energy}
\label{app:energy-snn}
Computing the energy requirement for an SNN follows the same principles used for ANNs. However, SNNs achieve energy savings via two properties of their processing: (a) not all neurons spike at each point in time, and (b) in the case of binary spikes (unit magnitude), matrix multiplications simplify to indexing and accumulation (addition) of elements of the weight matrices.
These savings become tangible during inference when deploying on specialized neuromorphic hardware \citep{Davies2018Loihi, true-north}.
Following \citep{zhu2024autonomous, qixu2023, mingqingOnlineSNN, li2023seenn}, we calculate these savings while training and running our models on GPUs.
To achieve this, a careful accounting is performed of the proportion of neurons that spiked at each spiking neuron layer for a given forward pass of a model.
That rate subsequently reduces the next layer's FLOP count.
For example, if the next layer is dense, the FLOP count for that layer is reduced proportionately to the spiking rate of the given layer.
Furthermore, as was explained earlier, ACs in hardware are cheaper than MACs.
This is another source of savings for SNNs, where each spike's forward computation is significantly cheaper than a whole floating-point MAC.

\subsubsection{Spiking MLPs}
\label{app:energy-smlps}
Here we describe the energy required per inference for a spiking MLP. Given a spiking MLP of L layers,
let $T$ denote the number of simulation steps used to update the state of each neuron.
\revisions{
To clarify, $T > 1$ values are commonly used in SNN works to signify that the same, time-independent input (e.g., a static image) is presented to the SNN $T$ times in succession, allowing neuron dynamics to converge.
The result of the network is then obtained by aggregation of the output layer's values over this artificially introduced time dimension.
In our case, $T=1$, since each sample of the input does not need to be repeated, due to our data's inherently temporal nature.
}
Furthermore, let $\rho_{l-1} \in [0,1]$ be the proportion of the previous layer's neurons that spiked, and $E_l$ be the energy consumed by the $l$-th layer.
Also let $E_W$ denote the energy per algebraic operation for a dense layer with weight matrix $\mathbf{W} \in \mathbb{R}^{n_i \times n_o}$ and bias vector $\mathbf{b} \in \mathbb{R}^{n_o}$, with $n_i, n_o$ being the input and output shapes respectively.

Then, for a batch size of 1, the equations for the energy consumption of a spiking MLP are as follows,
\begin{gather}
    E_{\textrm{SMLP}} = \sum_{l=1}^{L} T \cdot \rho_{l-1} \cdot n^{(l)}_i n^{(l)}_o \cdot E_{W} + n^{(l)}_o \cdot E_{\textrm{AC}},
    \label{eq:smlp-energy}
\end{gather}
where $l \in \{1, 2, \dots, N-1\}$ is the layer index.
This is similar to Eq. \ref{eq:mlp-energy}, the only differences being that there are spiking rates to help reduce the computational load, and that input values are binary spikes, meaning that: $E_W = E_{\textrm{AC}} < E_{\textrm{MAC}}$.
The latter holds because matrix-vector multiplication can be expressed purely in terms of indexing and accumulation when the vector is binary.

\subsubsection{Spiking self-attention}
We calculate spiking self-attention energy based on the formulation in App. \ref{app:ssa-definition}.
Let $L$ be the sequence length, $E$ the embedding dimension, $\mathbf{Q} \in \mathbb{R}^{L \times E}$ the query tensor, $\mathbf{K} \in \mathbb{R}^{L \times E}$ the key tensor, and $\mathbf{V} \in \mathbb{R}^{L \times D}$ the value tensor.
The computational requirements following Eq. \ref{eq:ssa}, for a single attention head, are computed as follows,
\begin{gather*}
    N_{\textrm{AC}}^{QK^\top} = L^2 E, ~
    N_{\textrm{MAC}}^{\textrm{norm}} = L^2, ~
    N_{\textrm{AC}}^{V} = L^2 D \Rightarrow \\
    N_{\textrm{AC}}^{\textrm{SSA}} = L^2 \cdot (E+D), ~\text{ and }
    N_{\textrm{MAC}}^{\textrm{SSA}} = L^2.
\end{gather*}
For \emph{multi-head attention}, we need only multiply the above by the number of heads.
Letting $H$ be the number of heads, the total energy for the cross attention operation is,
\begin{gather}
    E_\textrm{SSA} = H \cdot \left(
        E_{\textrm{MAC}} \cdot N_{\textrm{MAC}}^{\textrm{SSA}} + 
        E_{\textrm{AC}} \cdot N_{\textrm{AC}}^{\textrm{SSA}}
    \right).
    \label{eq:energy-ssa}
\end{gather}
Note that the cost for projecting inputs to $\mathbf{Q}, \mathbf{K}, \mathbf{V}$ and projecting outputs to the appropriate downstream shape is computed separately (spiking linear layers, computation follows the logic of \ref{app:energy-smlps}).

\subsubsection{Spikachu}
To compute the total energy consumed by Spikachu during a forward pass, we compute the sum of the energy of its modules, described below.
\paragraph{Harmonizer.} This is a CA block followed by a linear layer.
Its energy is computed following Eq. \ref{eq:energy-attn}, with the addition of the energy cost of the projection layers (denoted $E_h^{\textrm{Q}}, E_h^{\textrm{K}}, E_h^{\textrm{V}}, E_h^{\textrm{out}}$), and the extra linear layer (denoted $E_h^{\textrm{SMLP}}$),
\begin{gather}
    E_h =
        H \cdot E_{\textrm{MAC}} \cdot N_{\textrm{MAC}}^{\textrm{CA}} +
        E_h^{\textrm{Q}} +
        E_h^{\textrm{K}} +
        E_h^{\textrm{V}} +
        E_h^{\textrm{out}} +
        E_h^{\textrm{SMLP}},
\end{gather}
where each of the summands outside the parentheses is a direct application of Eq. \ref{eq:mlp-energy}.

\paragraph{Multi Scale SNN-I.} This is a combination of parallel spiking MLPs,  based on Eq. \ref{eq:smlp-energy},
\begin{gather}
    E_{s,1} = \sum_{p=1}^P E_{\textrm{SMLP}}^p
    = E_{\textrm{AC}} \cdot P \cdot
        \sum_{l=1}^{L}
            T \cdot
            \rho_{l-1} \cdot
                n^{(l)}_i n^{(l)}_o  + n^{(l)}_o,
\end{gather}
where $P$ is the number of parallel spiking MLPs and $l \in \{1, \dots, L\}$ is the layer index.

\paragraph{Spiking Self-Attention.} SSA is implemented as described in Sec. \ref{app:ssa-definition}, with energy computed based on Eq. \ref{eq:energy-ssa},
\begin{gather}
    E_a =
        H \cdot \left(
            E_{\textrm{MAC}} \cdot N_{\textrm{MAC}}^{\textrm{SSA}} + 
            E_{\textrm{AC}} \cdot N_{\textrm{AC}}^{\textrm{SSA}}
        \right)
    + E_\textrm{SSA}^{\textrm{Q}} +
        E_\textrm{SSA}^{\textrm{K}} +
        E_\textrm{SSA}^{\textrm{V}} +
        E_\textrm{SSA}^{\textrm{out}},
\end{gather}
where the added summands are spiking MLPs used for projections, their energy calculated using Eq. \ref{eq:smlp-energy}.

\paragraph{Spiking MLP.} Mixing module, energy given by Eq. \ref{eq:smlp-energy}, denoted $E_m$.

\paragraph{Multi Scale SNN-II.} Same as Multi Scale SNN---I, denoted $E_{s,2}$.

\paragraph{Membrane potential observer layer.} Layer containing only leaky integrators, energy per time-step is 1 AC operation per neuron,
\begin{gather}
    E_o = n_o \cdot E_\textrm{AC},
\end{gather}
where $n_o$ is the number of observer neurons.

\paragraph{Readout layer.} Single linear layer, equivalent to a single layer of Eq. \ref{eq:smlp-energy}, denoted by $E_r$.

Thus, the total energy per inference for Spikachu is given by,
\begin{gather}
    E_\textrm{Spikachu} = E_h + E_{s,1} + E_a + E_m + E_{s,2} + E_o + E_r.
\end{gather}

\newpage

\section{Additional Results} \label{app:additional_results}

\subsection{Scaling laws of multi-session, multi-subject training: Continued} \label{sec:scaling_laws_of_multi_subject_training_continued}

In Sec.~\ref{sec:scaling_laws}, we demonstrated that pretraining Spikachu on multi-session, multi-subject datasets improves both decoding performance and energy efficiency. Here, we present additional evidence that complements those findings.

\begin{figure}[h]
    \centering
    \includegraphics[scale=0.43]{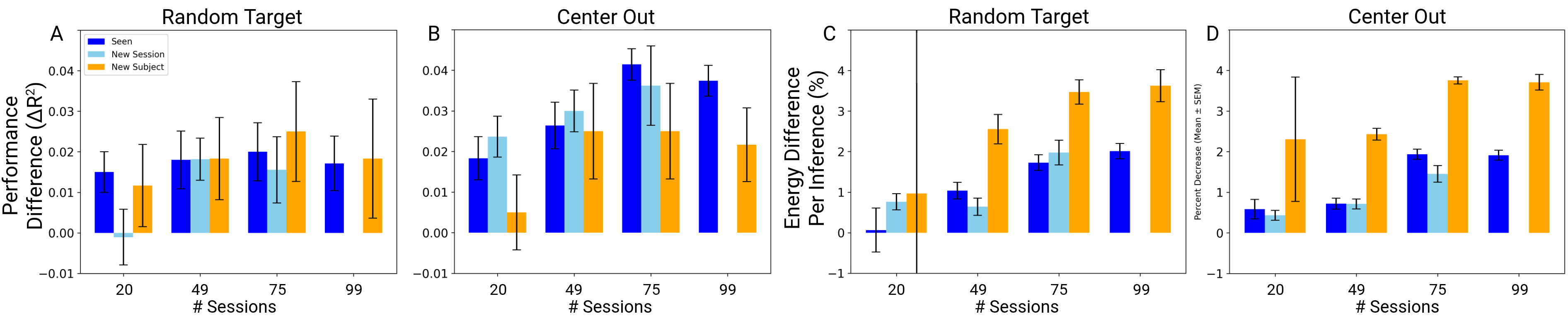} 
    \caption{\textit{Benefits of scaling up training dataset size.} \textbf{(A, B)} Performance difference between the single-subject models trained from scratch vs finetuned/transferred models from pretrained for the RT, and CO tasks. \textbf{(C, D)} Percent difference in energy consumption between the same groups for the same tasks.}
    \label{fig:performance_difference_when_scaling_up}
\end{figure}

Fig.~\ref{fig:performance_difference_when_scaling_up}A, B shows the improvement in per-session test set decoding performance achieved by the pretrained models after (1) finetuning on sessions seen during pretraining (monkeys C, J, and M; blue), (2) transferring to new sessions from animals seen during pretraining (monkeys C, J, and M; skyblue), and (3) transferring to new sessions from a new animal, unseen during pretraining (monkey T; orange). Across all conditions, we observe consistent performance gains that increase with the number of sessions used for pretraining, with gains possibly saturating beyond $\sim$75 sessions. %These results highlight the benefit of scaling pretraining to diverse, multi-subject data.

In Fig.~\ref{fig:performance_difference_when_scaling_up}C, D, we report the relative reduction in energy consumption per inference,
\[
\Delta E = -\frac{E_{\textrm{msf}} - E_{\textrm{ss}}}{E_{\textrm{ss}}},
\]
where $E_{\textrm{msf}}$ is the energy required per inference by the pretrained model after finetuning, and $E_{\textrm{ss}}$ is the energy used by a single-session model trained from scratch. Across all finetuning conditions, energy savings scale positively with the size of the pretraining dataset. Given that typical BCI sessions require tens of thousands of inferences per session, these efficiency gains are highly practical in a real-world setting.

\begin{figure}[h]
    \centering
    \includegraphics[scale=0.47]{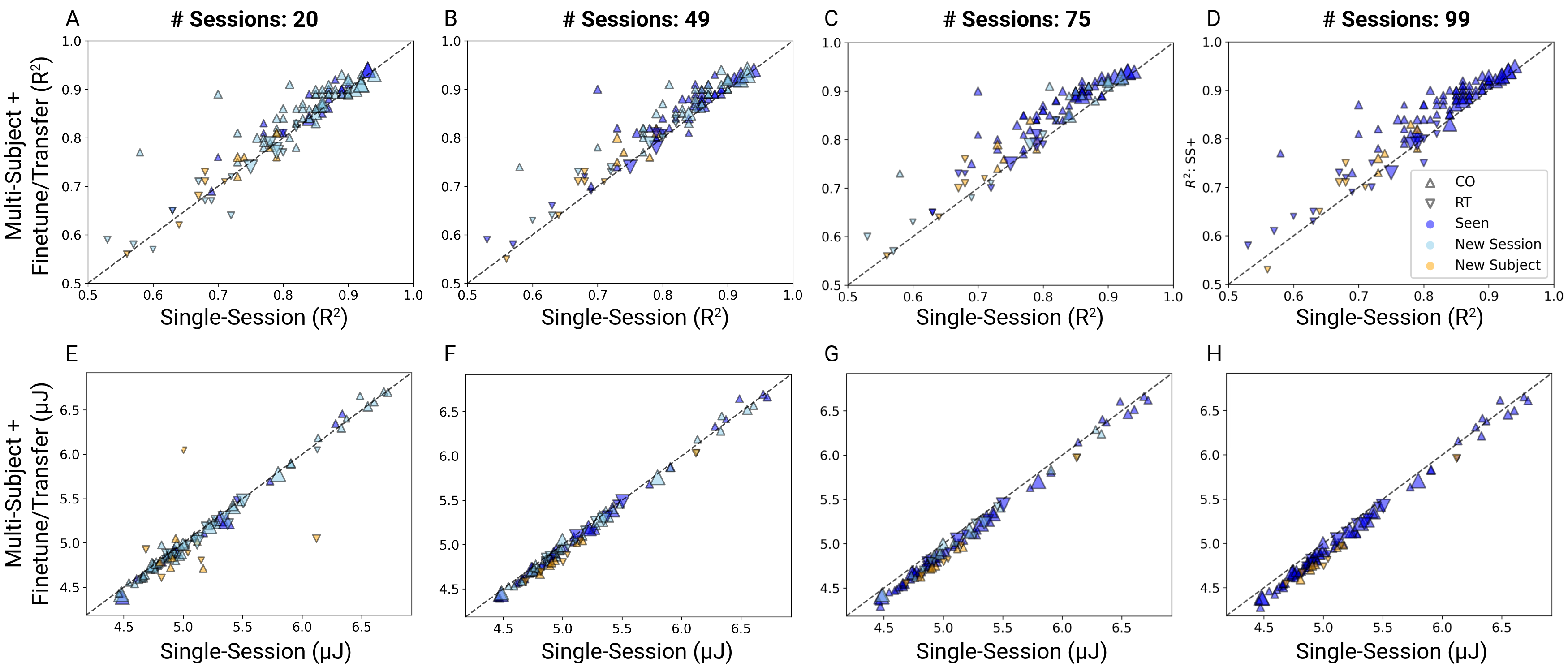} 
    \caption{\textit{Decoding performance and energy benefits of scaling up.}
    \textbf{(A, B, C, D)} Head-to-head decoding performance comparison between single-session models trained from scratch vs (1) the pretrained models + finetuning (seen; blue) and (2) pretrained models + transfer (new session; skyblue and new subject; orange).
    \textbf{(E, F, G, H)} Head-to-head comparison of the energy consumed per inference for the same groups.}
    \label{fig:scaling_analysis_performance}
\end{figure}

To further illustrate these benefits, Fig.~\ref{fig:scaling_analysis_performance}A--D presents head-to-head comparisons of decoding performance between finetuned models and their scratch-trained counterparts for pretraining on 20, 49, 75, and 99 sessions. In each case, the finetuned models outperform their baseline equivalents, with the performance gap widening as pretraining data increases. 
Figs.~\ref{fig:scaling_analysis_performance}E--H compare energy usage per inference under the same conditions. Here, we observe that finetuned models consistently consume less energy than their scratch-trained counterparts, emphasizing the energy efficiency achieved through pretraining. The energy savings also scale positively with the amount of pretraining data.

\begin{figure}[h]
    \centering
    \includegraphics[scale=0.47]{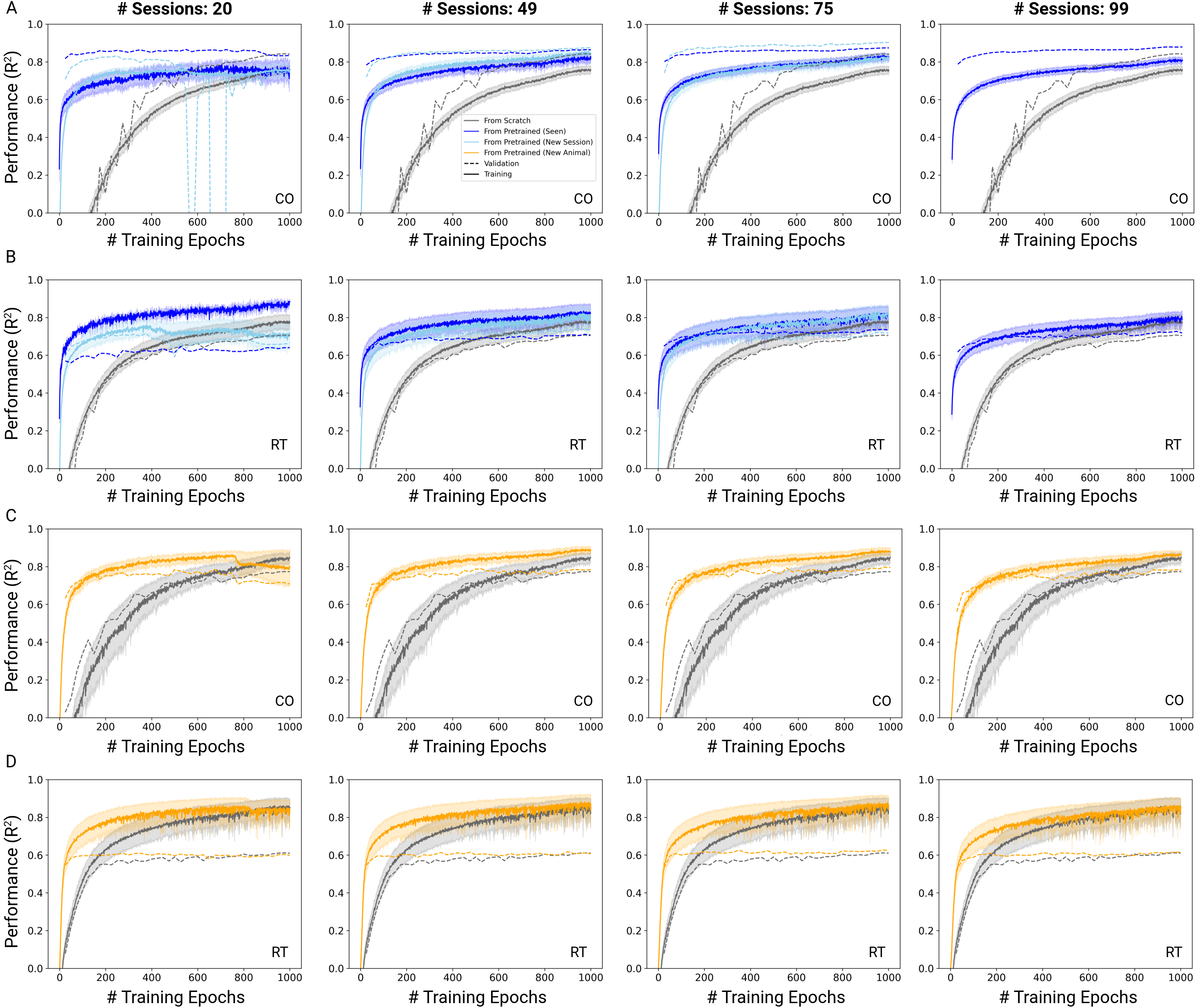} 
    \caption{\textit{Learning speedup achieved through pretraining}.   
    \textbf{(A)} Leaning Curves for the CO task for (1) single-session models trained from scratch (gray), (2) the pretrained models + finetuning (seen; blue), and (3) pretrained models + transfer (new session; skyblue).
    \textbf{(B)} Leaning Curves for the RT task for (1) single-session models trained from scratch (gray), (2) the pretrained models + finetuning (seen; blue), and (3) pretrained models + transfer (new session; skyblue).
    \textbf{(C)} Leaning Curves for the CO task for (1) single-session models trained from scratch (gray), and (2) pretrained models + transfer (new subject; orange).
    \textbf{(D)} Leaning Curves for the RT task for (1) single-session models trained from scratch (gray), and (2) pretrained models + transfer (new subject; orange).}
    \label{fig:scaling_analysis_learning_curves}
\end{figure}

Finally, Figs.~\ref{fig:scaling_analysis_learning_curves}A-D show the mean learning curves under the same three finetuning conditions, alongside the baseline learning curves of scratch-trained single-session models (gray). Across all levels of pretraining, finetuning the pretrained models to single sessions converges much faster than training single-session models from scratch. This demonstrates that the pretrained models learn transferable, general-purpose neural representations shared across both sessions and subjects that can be transferred very efficiently.

\newpage
\subsection{Ablation study} \label{app:ablation_study}

\subsubsection{Building blocks}

To identify the contribution of each building block of Spikachu to the decoding performance of our models, we performed an ablation study using the 99 recording sessions from monkeys C, J, M drawn from \citet{Perich2018}. Specifically, we trained variants of our proposed architecture, each time removing a building block and assessing the model's decoding performance. To ensure a fair comparison, training hyperparameters were kept identical across all training runs. 

The building blocks of our architecture are described in App. \ref{sec:spikachu_details} (see also Fig. \ref{networkArchitectrure}). For convenience, we also list them here: 
Harmonizer (Harm.),
Multi scale SNN-I (Multi-Scale I),
Spiking Self Attention (SSA),
Spiking MLP (sMLP),
Multi scale SNN-II (Multi-Scale II),
Membrane Potential Observer Layer,
Readout Layer.

We note that by ablating the multi-scale SNNs 
we used one spiking MLP instead of multiple parallel spiking MLPs. We did not ablate the sMLP 
because there is no straightforward way to project the output of the SSA block 
to the input dimensionality of Multi Scale SNN-II %(\textbf{[E]}) 
without it. For the same reason, we did not ablate the ``Readout Layer'' 
(needed to project the output of ``Membrane Potential Observer Layer'' %(\textbf{[E]}) 
to the output dimensionality of the cursor velocity). We also did not ablate the ``Membrane Potential Observer Layer'' %(\textbf{[F]}) 
because there is no straightforward way to track continuous variables (as is the velocity tracked in this work) without this layer when using SNNs.

\begin{figure}[h]
    \centering
    \includegraphics[scale=0.45]{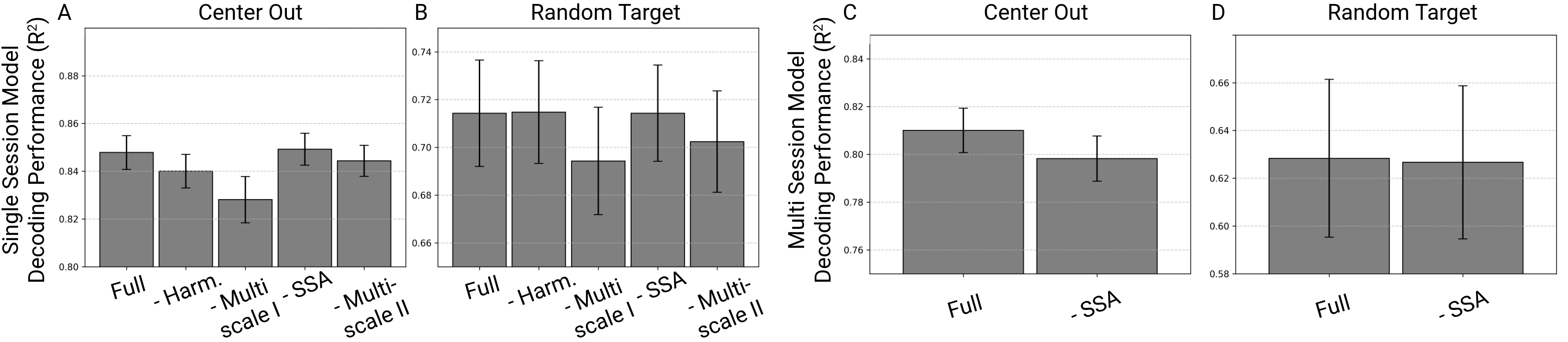} 
    \caption{\textit{Summary of ablation results}. \textbf{(A, B)} Single session training decoding performance (mean $\pm$ sem) in the validation set for the \textbf{(A)} CO and \textbf{(B)} RT tasks. 
    \textbf{(C, D)} Multi-session training decoding performance (mean $\pm$ sem) in the validation set for the \textbf{(A)} CO and \textbf{(B)} RT tasks for models trained on 75 sessions from 3 animals.}
    \label{fig:ablations.}
\end{figure}

The results of the ablation study when training Spikachu and variants in single sessions are shown in Fig. \ref{fig:ablations.}A, B. We observed that the spiking components of the architecture (``Multi Scale SNN-I'' and ``Multi Scale SNN-II'') influence model performance most, indicating that our model did not rely on the ANN part (the harmonizer) to perform.  We also observed that the performance difference when ablating the SSA block was negligible. To further investigate the utility of the SSA block, we trained variants of our architecture with the SSA block ablated on multiple sessions. We observed that the SSA block \underline{did} affect the model's performance when scaling up (see Fig. \ref{fig:ablations.}C, D for model performance on the validation set when training on 75 sessions).

\subsubsection{Spiking mechanism of neurons}

\begin{wraptable}{r}{0.6\textwidth}
\vspace{-4mm}
\centering
\caption{\footnotesize{\textit{Model performance for Spikachu and ANN Variants.}}}
\footnotesize
\begin{tabular}{l|rr}
\hline
\multirow{2}{*}{\textbf{Model Variant}} 
& \multicolumn{2}{c}{\textbf{Decod. Perf. (R\textsuperscript{2}) $\uparrow$}} \\
& \multicolumn{1}{c}{\textbf{CO}} & \multicolumn{1}{c}{\textbf{RT}} \\
\hline
ANN & 0.5332 & 0.3642 \\
ANN + context & 0.5348 & 0.3643 \\

SNN  & 0.8398 & 0.6761 \\
\hline
\end{tabular}

\label{tab:ANN_vs_SNN_ablations}
\vspace{-3mm}
\end{wraptable}

\revisions{To determine whether our model’s performance was driven by its spiking mechanism or simply by network connectivity, we conducted an additional ablation study. In this experiment, we re-implemented Spikachu as an ANN, preserving the exact same connectivity but replacing stateful LIF neurons with stateless ReLU neurons, thereby removing any intrinsic temporal dynamics. We refer to this variant as Spikachu-ANN.}

\revisions{We evaluated Spikachu-ANN on the same 99 recording sessions from monkeys C, J, and M from \citet{Perich2018} described in Section~\ref{sec:ss_training}, considering two different training conditions: (1) using only the current timebin (identical to the setup for Spikachu; ANN in Tab. \ref{tab:ANN_vs_SNN_ablations}), and (2) providing additional temporal context by including the four preceding timebins as inputs (ANN + context in Tab. \ref{tab:ANN_vs_SNN_ablations}).}

\revisions{As summarized in Tab. ~\ref{tab:ANN_vs_SNN_ablations}, Spikachu (SNN in Tab. \ref{tab:ANN_vs_SNN_ablations}) consistently outperformed its ANN counterpart under both conditions. Importantly, even when explicitly supplied with temporal context, Spikachu-ANN underperformed the spiking model. These findings demonstrate that Spikachu's performance arises from the LIF neuron's spiking dynamcis and is not due to network connectivity alone.}

\subsection{Testing the utility of the neural ``harmonizer'' on baseline models} \label{sec:homogen_with_baselines}

\begin{wraptable}{r}{0.6\textwidth}
\vspace{-4mm}
\centering
\caption{\footnotesize{\textit{Impact of harmonizer on model performance.}}}
\footnotesize
\begin{tabular}{l|cc|cc}
\hline
\multirow{2}{*}{\textbf{Model}} 
& \multicolumn{2}{c|}{\textbf{Center Out (R\textsuperscript{2}) }} 
& \multicolumn{2}{c}{\textbf{Random Target (R\textsuperscript{2})}} \\
& \multicolumn{1}{c}{\textbf{- Harm.}} & \multicolumn{1}{c|}{\textbf{+ Harm.}} 
& \multicolumn{1}{c}{\textbf{- Harm.}} & \multicolumn{1}{c}{\textbf{+ Harm.}} \\
\hline
LSTM & 0.4935 & 0.5804 & 0.4214 & 0.5919  \\
MLP & 0.7424 & 0.6415 & 0.5724 & 0.5229 \\
GRU & 0.8336 & 0.8187 & 0.6681 & 0.6110 \\
\hline
\end{tabular}

\label{tab:impact_of_harmonizer}
\vspace{-3mm}
\end{wraptable}

In this work, we introduce the neural harmonizer, a novel, causal method for aligning heterogeneous neural recordings across sessions and subjects, as detailed in Sec.~\ref{sec:harmonizer}. This approach addresses a key barrier to scaling neural decoding models to multi-session and multi-subject datasets: the reliance of traditional architectures, such as MLPs and GRUs, on homogeneous input structures and consistent neural correspondences. By projecting the disparate neural signals from different datasets into a unified representation, the harmonizer enables any standard model (see \citet{Glaser2020} for an overview) to be trained effectively across sessions and subjects, which could enhance the model's generalizability.

Although the homogenizer was developed as part of the Spikachu framework, we demonstrate its broader applicability by integrating it with baseline models (LSTM, MLP, and GRU) and training them across all 99 neural recording sessions from monkeys C, J, and M in the dataset from \citet{Perich2018}. The corresponding test set results, reported separately for the CO and RT tasks, are shown in Tab.~\ref{tab:impact_of_harmonizer}. 
The results indicate that the standard models perform well when paired with the neural harmonizer and even see performance gains in some cases.
We note that we did not perform any hyperparameter tuning for the harmonizer or the baseline models when combining them into a single pipeline. Instead, we used the harmonizer configuration optimized for Spikachu and the baseline models as trained in the single-session setting described in Sec.~\ref{sec:ss_training}.  This highlights the potential of our approach to scale model training across multi-session, multi-subject neural datasets, offering a flexible and powerful foundation for generalized neural decoding.  

\subsection{Representation analysis of the unit embedding space of \spikachump} \label{sec:representation_analysis_of_unit_embeddings}

\revisions{In this section, we investigated whether any structure emerged in the latent space of Spikachu-mp (see Sec. \ref{sec:ms_training}), the model we trained on 99 recording sessions from monkeys C, J, M performing the CO and RT tasks from \citet{Perich2018}. }

\begin{wrapfigure}{r}{0.6\textwidth} 
    \centering
    \includegraphics[scale=0.55]{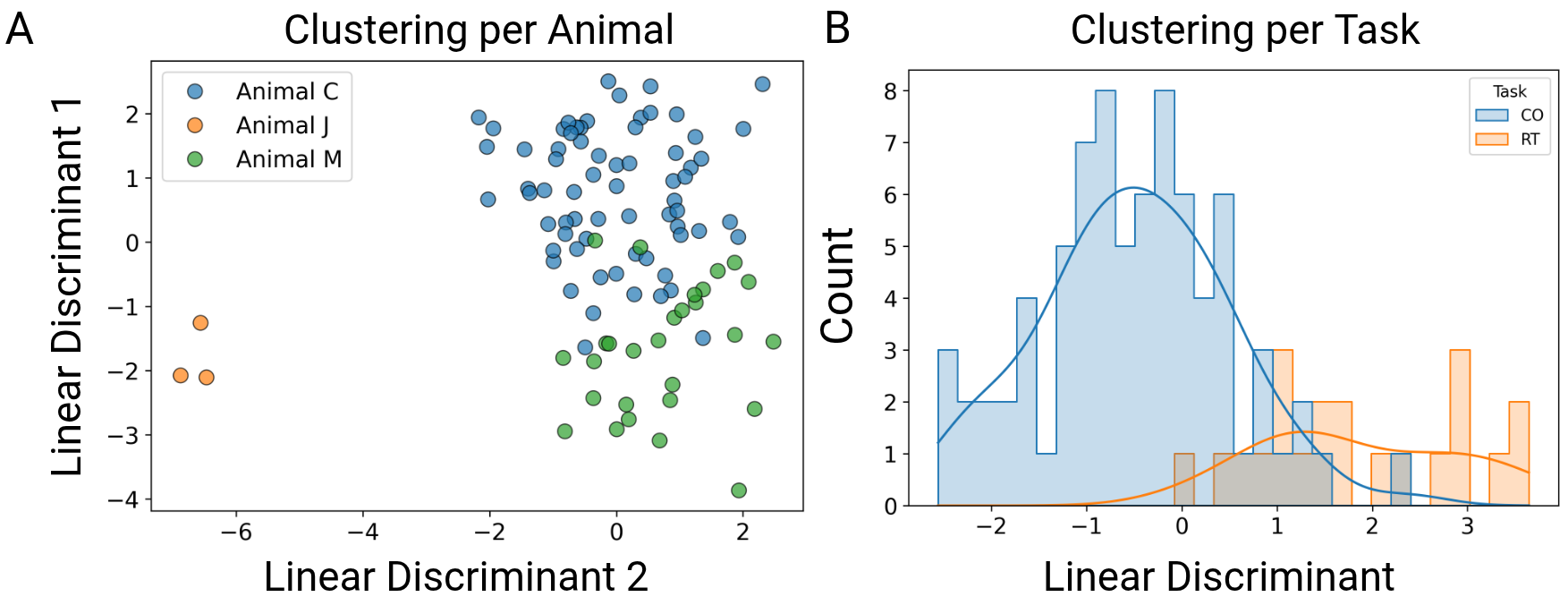}
    \caption{\textit{Linear Discriminant Analysis of session embeddings.} Visualization of session-level embeddings in the latent space when maximizing separability for \textbf{(A)} Subject, \textbf{(B)} Task.}
    \label{fig:LDAClustering}
\end{wrapfigure}

\revisions{Since our model does not have any explicit session-specific embeddings, for this analysis, we used the unit embeddings for each electrode of each session. To investigate whether meaningful structure emerged in the latent space of our trained model, we aggregated the unit embeddings within each session into a 2D matrix and applied PCA, retaining the first five principal components as a summary representation for each session. We then used these session-level representations as features in two distinct Linear Discriminant Analyses (LDA): one to assess separability by subject (monkeys C, J, and M) and another by task (Center Out vs. Random Target). Visualizing the session embeddings projected into the LDA space revealed clear clustering both by subject (see Fig. \ref{fig:LDAClustering}A) and by task (see Fig. \ref{fig:LDAClustering}B).}  

\revisions{To quantify the separability, we trained an SVM classifier (using scikit-learn’s SVC with default parameters) using 5-fold cross-validation with features the summary representations of each session (the first five principal components of the aggregated unit embeddings; same as those used in LDA) to predict (1) the subject and (2) the task associated with each session. We achieved an average accuracy of 0.69 ± 0.02 for subject classification (chance = 0.33) and 0.79 ± 0.02 for task classification (chance = 0.50), suggesting that \spikachump learned latent representations that reflect high-level structure in the data even though it is not explicitly trained to do so. }

\subsection{Profiling Spikachu in terms of memory access costs}
\label{sec:memory_ops}

\revisions{In Sec. \ref{sec:ss_training}, we benchmarked Spikachu's performance against various baseline models commonly used for neural decoding.
In this section, we provide comparisons between Spikachu and baselines in terms of computational and memory-access costs per inference. We note that this comparison is important since the total energy expenditure per inference consists of the energy required for computation, as well as the energy required to load (LOAD Ops) and store (STORE Ops) data in memory.}

\begin{wraptable}{r}{0.6\textwidth}
\vspace{-4mm}
\centering
\caption{\footnotesize{\textit{FLOP counts for Spikachu and baselines}. Best performing model is in bold and second best model is underlined. M stands for millions.}}
\footnotesize
\begin{tabular}{l|c|c|c}
\hline
\textbf{Model} & \textbf{MAC} (M) $\downarrow$ & \textbf{AC} (M) $\downarrow$ & \textbf{Total FLOPs} (M) $\downarrow$ \\
\hline
LSTM & 3.27 & 0.02 & 3.29 \\
MLP & 2.63 & 0.02 & 2.65 \\
POYO-causal & 466 & 1.0 & 467 \\
{GRU} & \underline{2.53} & \textbf{0.01} & \underline{2.54} \\
POYO & 466 & 1.0 & 467 \\
{Spikachu} & \textbf{0.97} & \underline{0.78} & \textbf{1.75} \\
\hline
\end{tabular}
\label{tab:flops}
\vspace{-3mm}
\end{wraptable}

\revisions{
For this analysis, we used the 99 single-session models trained on the recording sessions from monkeys C, J, and M performing the CO and RT tasks from \citet{Perich2018} (models produced by the experiment in Sec. \ref{sec:ss_training}). For each model and session, we calculated the number of FLOPs (i.e. MACS, and ACs) required for inference (see Sec. \ref{app:energy}).
Our findings for the number of necessary operations for each model are presented in Tab. \ref{tab:flops}.
Notably, Spikachu has a higher proportion of AC operations compared to other models, due to the binary output of each spiking neuron layer, which simplifies downstream computation.
\textit{In conclusion, Spikachu requires the least amount of such operations, thanks to its SNN backbone.}
} \footnote{In terms of energy expenditure, AC is significantly cheaper than MAC \citep{horowitz2014}.}

\revisions{Next, we use the aforementioned results to estimate memory access requirements by converting FLOPs to LOAD and STORE operations. As described in \citet{Liao2022}, we assume $N^{\textrm{LOAD}}_{\textrm{MAC}} = 3$ LOAD and $N^{\textrm{STORE}}_{\textrm{MAC}} = 1$ STORE operation per MAC and $N^{\textrm{LOAD}}_{\textrm{AC}} = 2$ LOAD and $N^{\textrm{STORE}}_{\textrm{AC}} = 1$ STORE operation per AC.
Therefore, memory operations can be estimated as,
\begin{gather}
    N^{\textrm{LOAD}} = N^{\textrm{LOAD}}_{\textrm{MAC}} \cdot N_{\textrm{MAC}} + N^{\textrm{LOAD}}_{\textrm{AC}} \cdot N_{\textrm{AC}} \\
    N^{\textrm{STORE}} = N^{\textrm{STORE}}_{\textrm{MAC}} \cdot N_{\textrm{MAC}} + N^{\textrm{STORE}}_{\textrm{AC}} \cdot N_{\textrm{AC}}
\end{gather}
where $N_{\textrm{AC}}$ and $ N_{\textrm{MAC}}$ denote the total number of AC and MAC operations for a given architecture, respectively.
}

\begin{wraptable}{r}{0.6\textwidth}
\vspace{-4mm}
\centering
\caption{\footnotesize{\textit{Memory access counts for Spikachu and baselines}. Best performing model is in bold and second best model is underlined. M stands for millions.}}
\footnotesize
\begin{tabular}{l|c|c|c}
\hline
\textbf{Model} & \textbf{Load} (M) $\downarrow$ & \textbf{Store} (M) $\downarrow$ & \textbf{Total Ops} (M) $\downarrow$ \\
\hline
LSTM & 9.85 & 3.29 & 13.14 \\
MLP & 7.93 & 2.65 & 10.58 \\
POYO-causal & 1400 & 467 & 1860 \\
\underline{GRU} & \underline{7.61} & \underline{2.54} & \underline{10.15} \\
POYO & 1400 & 467  & 1860 \\
\textbf{Spikachu} & \textbf{4.47} & \textbf{1.75} & \textbf{6.22} \\
\hline
\end{tabular}
\label{tab:memory-reference}
\vspace{-3mm}
\end{wraptable}

\revisions{
The results, averaged across all sessions and tasks for Spikachu and baselines, are summarized in Tab. \ref{tab:memory-reference}. \emph{Notably, Spikachu required the fewest memory access costs across the board, highlighting its energy efficiency not only from reduced computational demands but also from minimized memory costs}. This result further strengthens the promise of energy-efficient deployment of Spikachu for fully implantable BCIs.}

\revisions{We also note that while this analysis is hardware-agnostic, the practical energy savings are likely substantially greater than what is suggested here. The present analysis is most directly applicable to von Neumann architectures (e.g., conventional CPUs and GPUs), where further memory overheads arise from maintaining DRAM blocks even when their contents are not accessed during a clock cycle \citep{vonNeuman1003first}. In contrast, neuromorphic hardware (the intended deployment platform for Spikachu) typically relies on local SRAM memory, which incurs negligible keep-alive costs and only needs to supply data on a per-core basis. This architecture effectively eliminates the bulk of memory-related costs compared to von Neumann systems. As a result, energy savings are expected to be significantly more pronounced on neuromorphic hardware, with total memory costs likely falling below the estimates derived from our hardware-agnostic analysis presented in this section.}

\newpage
\section*{NeurIPS Paper Checklist}

\begin{enumerate}

\item {\bf Claims}
    \item[] Question: Do the main claims made in the abstract and introduction accurately reflect the paper's contributions and scope?
    \item[] Answer: \answerYes{} % Replace by \answerYes{}, \answerNo{}, or \answerNA{}.
    \item[] Justification:
        To the best of our knowledge, all claims made in the abstract and introduction are supported by our experiments described in Sec. \ref{sec:experiments} and App. \ref{app:additional_results}.

\item {\bf Limitations}
    \item[] Question: Does the paper discuss the limitations of the work performed by the authors?
    \item[] Answer: \answerYes{} % Replace by \answerYes{}, \answerNo{}, or \answerNA{}.
    \item[] Justification:
    In Sec. \ref{sec:discussion}, we discuss the limitations of this work.

\item {\bf Theory assumptions and proofs}
    \item[] Question: For each theoretical result, does the paper provide the full set of assumptions and a complete (and correct) proof?
    \item[] Answer: \answerYes{} % Replace by \answerYes{}, \answerNo{}, or \answerNA{}.
    \item[] Justification:
    The only theoretical results in this work describe our process for calculating energy requirements for various machine learning models in App. \ref{app:energy}. To the best of our knowledge, they are complete and correct.

    \item {\bf Experimental result reproducibility}
    \item[] Question: Does the paper fully disclose all the information needed to reproduce the main experimental results of the paper to the extent that it affects the main claims and/or conclusions of the paper (regardless of whether the code and data are provided or not)?
    \item[] Answer: \answerYes{} % Replace by \answerYes{}, \answerNo{}, or \answerNA{}.
    \item[] Justification:
        Detailed information required to reproduce all experiments can be found in Sec. \ref{sec:experiments} and App. \ref{app:model_implementation_details}, \ref{app:spikachu-training}, and \ref{app:energy}. 

\item {\bf Open access to data and code}
    \item[] Question: Does the paper provide open access to the data and code, with sufficient instructions to faithfully reproduce the main experimental results, as described in supplemental material?
    \item[] Answer: \answerYes{} % Replace by \answerYes{}, \answerNo{}, or \answerNA{}.
    \item[] Justification:
        The dataset used in this work has been made publically available by \citet{Azabou2023} here: \\ \texttt{https://github.com/neuro-galaxy/poyo}.
        We will make our code publically available upon acceptance of this manuscript.

\item {\bf Experimental setting/details}
    \item[] Question: Does the paper specify all the training and test details (e.g., data splits, hyperparameters, how they were chosen, type of optimizer, etc.) necessary to understand the results?
    \item[] Answer: \answerYes{} % Replace by \answerYes{}, \answerNo{}, or \answerNA{}.
    \item[] Justification:
        All details of our training can be found in App. \ref{app:spikachu-training}.% in the appendix, including a full list of hyperparameters and the type of optimizer.
        We also share the data splits and random seed used to generate them in App. \ref{appendix:sec:data}.

\item {\bf Experiment statistical significance}
    \item[] Question: Does the paper report error bars suitably and correctly defined or other appropriate information about the statistical significance of the experiments?
    \item[] Answer: \answerYes{} % Replace by \answerYes{}, \answerNo{}, or \answerNA{}.
    \item[] Justification:
        Error bars report mean $\pm$ standard error of the mean for all plots, unless otherwise mentioned. No experiments to assess statistical significance were performed in this work.

\item {\bf Experiments compute resources}
    \item[] Question: For each experiment, does the paper provide sufficient information on the computer resources (type of compute workers, memory, time of execution) needed to reproduce the experiments?
    \item[] Answer: \answerYes{} % Replace by \answerYes{}, \answerNo{}, or \answerNA{}.
    \item[] Justification:
        App. \ref{app:compute} describes the compute resources used in this work. %is dedicated to computational requirements (accelerator types, memory, time) used for our experiments.
    
\item {\bf Code of ethics}
    \item[] Question: Does the research conducted in the paper conform, in every respect, with the NeurIPS Code of Ethics \url{https://neurips.cc/public/EthicsGuidelines}?
    \item[] Answer: \answerYes{} % Replace by \answerYes{}, \answerNo{}, or \answerNA{}.
    \item[] Justification:
        To the best of our knowledge, this work complies in every aspect with the NeurIPS Code of Ethics. %It does not involve human subjects, personal data, or sensitive information, and all datasets used are publicly available and properly cited, respecting their licenses and terms of use. The proposed methods do not promote or facilitate harmful, discriminatory, deceptive, or illegal activities, nor do they create environmental risks or security vulnerabilities. Potential societal impact is considered minimal and positive, as the work focuses on advancing energy-efficient, causal, and generalizable decoding methods in neuroscience applications, which aim to improve accessibility and assistive technologies while maintaining responsible research practices.

\item {\bf Broader impacts}
    \item[] Question: Does the paper discuss both potential positive societal impacts and negative societal impacts of the work performed?
    \item[] Answer: \answerYes{} % Replace by \answerYes{}, \answerNo{}, or \answerNA{}.
    \item[] Justification:
        Please see Sec. \ref{sec:intro} and \ref{sec:discussion}.
        We have not discussed negative societal impacts, since there are none to the best of our knowledge.
    
\item {\bf Safeguards}
    \item[] Question: Does the paper describe safeguards that have been put in place for responsible release of data or models that have a high risk for misuse (e.g., pretrained language models, image generators, or scraped datasets)?
    \item[] Answer: \answerNA{} % Replace by \answerYes{}, \answerNo{}, or \answerNA{}.
    \item[] Justification:
        We have not discussed such safeguards, as there is no foreseeable potential for misuse of our work.
        We hope that the community understands the motivation behind our approach and applies it responsibly to improve the lives of the intended patients.

\item {\bf Licenses for existing assets}
    \item[] Question: Are the creators or original owners of assets (e.g., code, data, models), used in the paper, properly credited and are the license and terms of use explicitly mentioned and properly respected?
    \item[] Answer: \answerYes{} % Replace by \answerYes{}, \answerNo{}, or \answerNA{}.
    \item[] Justification:
        All assets used for our work are licensed for academic use.
        We also have an extensive list of citations, ensuring proper attribution of the works of others we used in this manuscript.
        Where applicable, we have acquired explicit permission to reproduce explanatory graphics in our figures.

\item {\bf New assets}
    \item[] Question: Are new assets introduced in the paper well documented and is the documentation provided alongside the assets?
    \item[] Answer: \answerYes{} % Replace by \answerYes{}, \answerNo{}, or \answerNA{}.
    \item[] Justification:
        Our code is well documented and will be made publically available upon acceptance of this manuscript.
        There are no other assets generated by this work that require documentation.

\item {\bf Crowdsourcing and research with human subjects}
    \item[] Question: For crowdsourcing experiments and research with human subjects, does the paper include the full text of instructions given to participants and screenshots, if applicable, as well as details about compensation (if any)? 
    \item[] Answer: \answerNA{} % Replace by \answerYes{}, \answerNo{}, or \answerNA{}.
    \item[] Justification:
        This work does not involve any research with human subjects.

\item {\bf Institutional review board (IRB) approvals or equivalent for research with human subjects}
    \item[] Question: Does the paper describe potential risks incurred by study participants, whether such risks were disclosed to the subjects, and whether Institutional Review Board (IRB) approvals (or an equivalent approval/review based on the requirements of your country or institution) were obtained?
    \item[] Answer: \answerNA{} % Replace by \answerYes{}, \answerNo{}, or \answerNA{}.
    \item[] Justification:
        This work does not involve any research with human subjects.

\item {\bf Declaration of LLM usage}
    \item[] Question: Does the paper describe the usage of LLMs if it is an important, original, or non-standard component of the core methods in this research? Note that if the LLM is used only for writing, editing, or formatting purposes and does not impact the core methodology, scientific rigorousness, or originality of the research, declaration is not required.
    \item[] Answer: \answerNA{} % Replace by \answerYes{}, \answerNo{}, or \answerNA{}.
    \item[] Justification:
        LLMs did not contribute to the research in this work.

\end{enumerate}

\end{document}